\documentclass{bmvc2k}

\usepackage{graphicx}
\usepackage{booktabs}
\usepackage{makecell}
\usepackage{wrapfig}
\usepackage{graphicx}
\usepackage{verbatim}
\usepackage{amsmath}
\usepackage[capitalize]{cleveref}

\usepackage{colortbl} 
\usepackage{xcolor} 
\usepackage{arydshln}

\usepackage[percent]{overpic}
\usepackage{pifont}
\usepackage{algorithm}
\usepackage{algpseudocode}
\usepackage{multirow}
\usepackage{float}
\usepackage{array}
\usepackage{tikz}
\usepackage{bm}
\usepackage{bbding}
\usepackage{pifont}
\usepackage{wasysym}
\usepackage{amssymb}
\usepackage{pifont}
\usepackage{tabularx}
\usepackage{booktabs}
\usepackage[export]{adjustbox}
\usepackage{enumitem}
\usepackage[accsupp]{axessibility}  
\usepackage{url}

\definecolor{First}{HTML}{BDE6CD}
\definecolor{Second}{HTML}{E2EEBC}
\definecolor{Third}{HTML}{FFF8C5}

\newcommand{\fst}[1]{\cellcolor{First}#1}
\newcommand{\snd}[1]{\cellcolor{Second}#1}
\newcommand{\trd}[1]{\cellcolor{Third}#1}

\title{Self-Evolving Depth-Supervised 3D Gaussian Splatting from Rendered Stereo Pairs}

\addauthor{Sadra Safadoust}{https://sadrasafa.github.io/}{1*}
\addauthor{Fabio Tosi}{https://fabiotosi92.github.io/}{2}
\addauthor{Fatma Güney}{https://mysite.ku.edu.tr/fguney/}{1}
\addauthor{Matteo Poggi}{https://mattpoggi.github.io/}{2}
\addinstitution{
 Department of Computer Engineering \\
 and KUIS AI Center, 
 Koç University, \\
 Istanbul, Turkey
}
\addinstitution{
 Department of Computer Science and Engineering (DISI),\\
 University of Bologna, 
 Italy \\
 \vspace{0.3cm}
 \scriptsize{* project started while visiting University of Bologna}
}

\runninghead{S. Safadoust et al.}{Self-Evolving 3D Gaussian Splatting}

\begin{document}

\maketitle

\begin{abstract}
3D Gaussian Splatting (GS) significantly struggles to accurately represent the underlying 3D scene geometry, resulting in inaccuracies and floating artifacts when rendering depth maps. In this paper, we address this limitation, undertaking a comprehensive analysis of the integration of depth priors throughout the optimization process of Gaussian primitives, and present a novel strategy for this purpose. This latter dynamically exploits depth cues from a readily available stereo network, processing virtual stereo pairs rendered by the GS model itself during training and achieving consistent self-improvement of the scene representation. Experimental results on three popular datasets, breaking ground as the first to assess depth accuracy for these models, validate our findings. 
\end{abstract}

{\hfill  \small Project page: \url{https://kuis-ai.github.io/StereoGS/} \hfill }

\begin{figure}[h]
    \centering
    \includegraphics[width=0.95\textwidth]{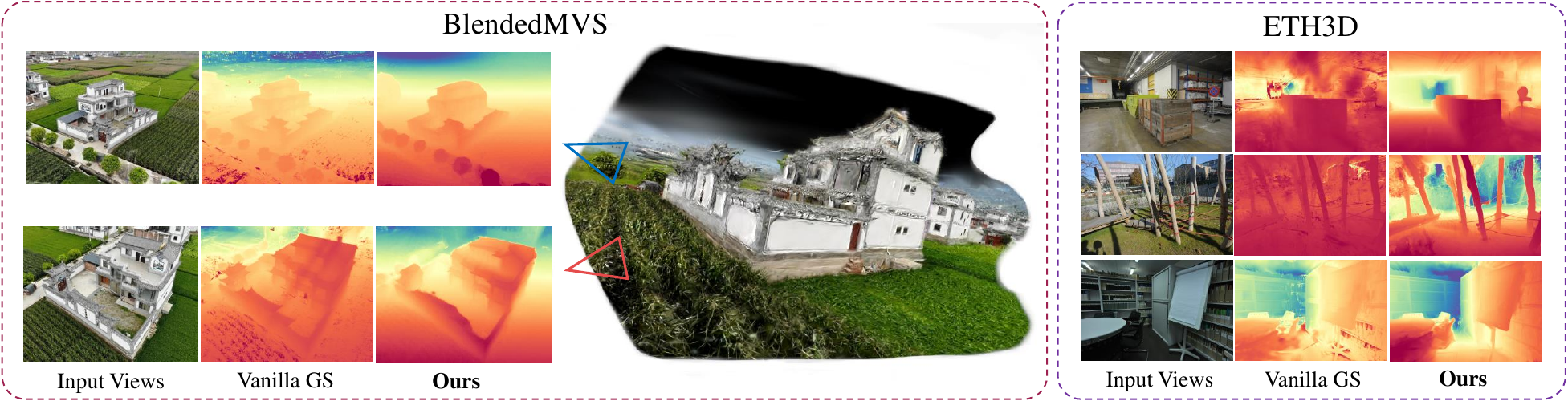}
   \vspace{-0.3cm}
   \caption{\textbf{Self-Evolving Depth-Supervised 3D Gaussian Splatting (GS) in action.} 
   Our strategy allows GS to self-improve during optimization, and to render better depth maps.
}\label{fig:teaser}\vspace{-0.3cm}
\end{figure}

\section{Introduction}
\label{sec:intro}

In recent years, NeRF \cite{mildenhall2020nerf} has deeply revolutionized several aspects of computer vision, introducing innovative paradigms and redefining our understanding of the field. First and foremost, NeRF has represented a turning point for image rendering and novel view synthesis \cite{mildenhall2020nerf,zhang2020nerf++,martin2021nerf}, casting these tasks as the optimization of a continuous 3D representation of the scene encoded in multi-layer perceptrons (MLPs), queried with $(x,y,z)$ coordinates in space and $(\theta,\phi)$ view angles to predict color and opacity for any generic 3D point. Rendering is achieved by casting rays and accumulating colors and opacities along them into pixels.

However, the recent advent of 3D Gaussian Splatting (GS) \cite{kerbl20233d} is rapidly conquering the main stage at the expense of NeRF, due to the much lower time required for both optimization and rendering.
While GS has certainly advanced the state-of-the-art in terms of photorealism and rendering speed, the underlying geometry modeled by the Gaussians does not reflect the same quality as the rendering. This is evident in the depth maps generated by GS itself -- by replacing the color components of the Gaussians with their position during the rendering process -- as we can notice in \cref{fig:teaser}, where examples from BlendedMVS and ETH3D are affected by several floaters and artifacts.
Some works from the literature \cite{attal2021torf,kangle2021dsnerf,roessle2022dense,xu2022sinnerf,wang2023digging} show evidence that using depth priors as a form of additional supervision when optimizing a NeRF can improve the quality of the rendered images, especially when very few images are available for training. Intuitively, this strategy also has the potential to improve the underlying geometry encoded by the NeRF itself, although no attention has been paid to this by prior works, nor to the different, possible approaches for retrieving depth priors from the very same images used to train NeRF -- and to the intrinsic limitation each choice brings. 

In this paper, we first delve into a study of Depth-Supervised 3D Gaussian Splatting (DS-GS) variants by examining and measuring the impact that different depth-from-image solutions have on the optimization of both the appearance and geometry modeled by GS. Specifically, 
we review four main strategies for extracting depth priors from the multiple images involved in GS optimization: i) Structure-from-Motion (SfM) \cite{schoenberger2016mvs}, ii) Monocular Depth Estimation (MDE) \cite{bhat2023zoedepth}, iii) Depth Completion (DC) \cite{bartolomei2023revisiting}, and iv) Multi-View Stereo (MVS) \cite{ma2022multiview}. Each one has its peculiar strengths, as well as its weaknesses. 
To name a few: on the one hand, 
SfM and MVS suffer in the presence of sparse views where the overlap between images is small; on the other hand, single-image approaches are free from this constraint, yet assume a network that can generalize properly across very different scenarios.

In addition to this exploration, we propose a novel approach to improve GS optimization, still by exploiting the supervision of depth priors, this time obtained by a fifth strategy that is a cornerstone of computer vision, but not included in the previous list: \textit{stereo matching}.
Indeed, we argue that despite the inaccurate underlying geometry modeled by the Gaussians, GS can still render geometrically consistent images -- e.g., rectified stereo images, even when a stereo camera is unavailable. 
Accordingly, GS can be employed to generate frames, which can then be processed by a pre-trained deep stereo network \cite{lipson2021raft} to obtain the supplementary supervision required to enhance the underlying geometry of GS itself. Thanks to the efficient rasterization process of GS, we can carry this out directly during the optimization, deploying a new GS framework capable of \textit{self-evolving}, supported by this external stereo network.

Experiments on ETH3D \cite{schops2017multi}, ScanNet++ \cite{yeshwanthliu2023scannetpp} and BlendedMVS \cite{yao2020blendedmvs} support our claims:

    $\bullet$ We carry out a comparison between different strategies for retrieving depth priors from images, by evaluating the impact of each on improving both the appearance and geometry modeled by DS-GS against vanilla GS.
    
    $\bullet$ We propose a new Self-Evolving GS framework, capable of supervising itself through depth priors retrieved by a stereo matching network, processing the rectified images rendered by the GS itself during training. 

    $\bullet$ Compared to the use of depth priors from classical strategies, our approach renders both better images and depth maps in real, sparse view settings.

\section{Related Work}
\label{sec:related_work}

In this section, we present a review of the literature relevant to our study.

\textbf{Novel View Synthesis}. Generating realistic novel views from 3D scene representations has been an active area of research. Early approaches used geometries such as meshes \cite{hu2021worldsheet, riegler2020free, riegler2021stable}, planes \cite{hoiem2005automatic}, and point clouds \cite{xu2022point, zhang2022differentiable} to model scenes. More recently, neural radiance fields (NeRFs) \cite{mildenhall2020nerf}, extensively discussed in \cite{rabby2023beyondpixels, dellaert2020neural, gao2022nerf}, have emerged as a highly effective representation for photorealistic novel view synthesis. NeRF models the scene as an implicit, continuous radiance field, allowing fine-grained detail to be captured. Several extensions to the original NeRF formulation have enhanced rendering quality by improving anti-aliasing \cite{barron2023zip, barron2022mip, barron2021mip}, modeling reflectance more accurately \cite{guo2022nerfren, chen2023nerrf, verbin2022ref}, training with sparse views \cite{niemeyer2022regnerf,xu2022sinnerf,kim2022infonerf}, and reducing computational overhead during training \cite{mueller2022instant, reiser2021kilonerf, hu2022efficientnerf, sun2022direct, fridovich2022plenoxels} and rendering time \cite{garbin2021fastnerf,liu2020neural, yu2021plenoctrees}. 
In parallel, point cloud-based representations \cite{xu2022point, yifan2019differentiable, zhang2022differentiable} have gained popularity due to their rendering efficiency. In addition, recent differentiable point splatting techniques, such as 3D GS \cite{kerbl20233d}, have enabled state-of-the-art real-time scene rendering.
Among others, NeRF was used to generate stereo images for training stereo networks \cite{Tosi_2023_CVPR}. Our work has a different goal -- to generate stereo pairs and supervise GS itself. 

\textbf{Image-based Depth Estimation}. Traditionally, depth estimation from images has relied on non-learning-based approaches. Among them, Structure-from-Motion \cite{agarwal2009building,gherardi2010improving,schoenberger2016sfm} 
estimates both the sparse 3D structure of a scene and the camera poses from a set of images, with COLMAP \cite{schoenberger2016sfm} 
becoming a reference pipeline for the community in the last decade.
When poses are known already, Multi-View Stereo \cite{furukawa2015multi, seitz2006comparison} used 
feature matching and geometric constraints -- i.e., epipolar geometry -- across multiple views, while binocular stereo \cite{scharstein2002taxonomy, hamzah2016literature} relied on correspondences between rectified stereo pairs. 
However, the field has undergone a revolutionary transformation with the advent of deep learning. In the context of multi-view \cite{yan2021deep, stathopoulou2023survey, wang2021multi} and binocular stereo \cite{poggi2021synergies, hamid2022stereo, zhou2020review, tosi2024survey}, learning-based approaches have leveraged the capabilities of convolutional neural networks to extract robust feature representations for more accurate correspondence estimation. This infusion of deep learning has not only increased accuracy, but has also facilitated the refinement and completion of sparse/noisy depth maps \cite{hu2022deep, khan2022comprehensive, xie2022recent, hu2022deep}, effectively filling in holes or refining inaccuracies. Within this paradigm, monocular depth estimation \cite{zhao2020monocular, bhoi2019monocular, masoumian2022monocular, ming2021deep} has emerged as a specialized subset, where deep neural networks are trained to directly predict depth out of a single image, typically 
on large datasets where pseudo-ground truth depth is available \cite{Ranftl2021, Ranftl2022}.

\textbf{Radiance Fields with Depth Priors}. While successful in image rendering, challenges arise in representing accurate scene geometries using advances in radiance fields. In response, supervised approaches incorporating depth priors have recently emerged.  
As a precursor, DS-NeRF \cite{deng2022depth} employs depth supervision using sparse point clouds from COLMAP during training. 
In parallel, Rossle et al. \cite{roessle2022dense} employ dense depth priors by densifying sparse depth data extracted from COLMAP. 
In NerfingMVS \cite{wei2021nerfingmvs}, instead, COLMAP extracts sparse depth priors, subsequently utilized to fine-tune a pretrained monocular depth network that is then 
employed to supervise volume sampling. 
PointNeRF \cite{xu2022point} introduces an intermediate step using feature point clouds and demonstrates improved efficiency compared to the vanilla NeRF. To the same end, CorresNeRF \cite{lao2024corresnerf} uses adaptive correspondence generation, while MonoSDF \cite{yu2022monosdf} improves the reconstruction process by incorporating depth and normal cues predicted by general-purpose monocular estimators. Similarly, SparseNeRF \cite{wang2023sparsenerf} leverages depth priors from real-world inaccurate observations, which can be from pre-trained depth models or coarse depth maps of consumer-level depth sensors, 
while some approaches used depth priors when dealing with dynamic scenes \cite{gerats2023dynamic, gao2021dynamic}. 
In a concurrent effort, Chung et al. \cite{chung2023depth} represents the only attempt to 
regularize GS using monocular depth networks. However, our paper highlights that better priors can be exploited to this end.

\section{Background Theory}
\label{sec:background}

\subsection{3D Gaussian Splatting (GS)} \label{sec:GS} 

3D Gaussian Splatting is a groundbreaking technique in the domain of explicit radiance fields and computer graphics. This unique approach relies on the incorporation of millions of 3D Gaussians, which marks a shift from the prevailing methods used in neural radiance field. 

Learning starts with multi-view images, by estimating camera poses and (optionally) sparse 3D points to bootstrap GS, which optimizes a set $\mathcal{G}=\{g_1, g_2, \dots, g_\mathcal{N}\}$ of 3D Gaussians, where $\mathcal{N}$ is the number of Gaussians in the scene. Each Gaussian, denoted as $g_i$, is characterized by a full 3D covariance matrix \({\Sigma}_i \in \mathbb{R}^{3 \times 3}\), center position \(\bm{\mu}_i \in \mathbb{R}^3\), opacity \(o_i \in [0,1]\), and color $\mathbf{c}_i$, which is represented by spherical harmonics (SH) for a view-dependent appearance. Backpropagation can be used to learn and optimize all these properties. The spatial influence of a single Gaussian primitive can be expressed as follows:

\small\begin{equation}
    g_i(\mathbf{x}) = e^{-\frac{1}{2} (\mathbf{x}-\bm{\mu}_i)^\top {\Sigma}_i^{-1} (\mathbf{x}-\bm{\mu}_i)}
\end{equation}\normalsize

Here, the spatial covariance ${\Sigma}$ defines an ellipsoid as ${\Sigma} = \mathbf{RSS^{\top}R^{\top}}$, where $\mathbf{S} \in \mathbb{R}^3$ represents the spatial scale and $\mathbf{R} \in \mathbb{R}^{3 \times 3}$ represents the rotation, parameterized by a quaternion.

For rendering, GS operates similarly to NeRF but deviates significantly in the computation of blending coefficients. 
This involves the "splatting" of 3D Gaussian points onto a 2D image plane, as ${\Sigma}' = \mathbf{JW} \Sigma \mathbf{W^{\top} J^{\top}}$ and $\bm{\mu}' = \mathbf{JW}\bm{\mu}$. Then, pixel color $C$ is obtained by merging 3D Gaussian splats that overlap, sorted by depth:

\small\begin{equation}
C = \sum_{i \in \mathcal{N}} \mathbf{c}_i \alpha_i \prod_{j=1}^{i-1} (1 - \alpha_j) \quad\quad\quad \text{with} \quad\quad \alpha_i = o_i \exp\left(-\frac{1}{2} (\mathbf{x}' - \bm{\mu}'_i)^\top {\Sigma}'^{-1}_i (\mathbf{x}' - \bm{\mu}'_i)\right)
\end{equation}\normalsize
The optimization process begins 
either from Structure-from-Motion (SfM) point clouds or random 3D points.
Then, Stochastic Gradient Descent (SGD) is employed, with L1 and D-SSIM loss functions between real and rendered views. 
Analogously to color, the resulting depth $D$ can be determined by replacing $\mathbf{c}_i$ with $d_i$ -- i.e., the distance of $g_i$ from the camera.

\subsection{Depth from Images}
\label{sec:depth}

In this section, we review established methodologies for estimating depth from images tailored for GS setting 
-- i.e., multiple images captured from different viewpoints using a single camera -- allowing to implement Depth-Supervised 3D Gaussian Splatting (DS-GS) variants. 

\textbf{Structure-from-Motion (SfM).}
It aims at reconstructing 3D structure and camera positions from a set of images, starting with two-view triangulation:

\small\begin{equation}
    \quad\quad \mathbf{X}_{ij} \sim \tau (\Tilde{\mathbf{x}}_i, \Tilde{\mathbf{x}}_j, \mathbf{P}_i, \mathbf{P}_j) \quad\quad \text{with} \quad i \neq j
\end{equation}\normalsize
with $\mathbf{X}_{ij}$ being a generic 3D point visible from images $\mathbf{I}_i, \mathbf{I}_j$, $\Tilde{\mathbf{x}}_i, \Tilde{\mathbf{x}}_j$ its pixel coordinates on the two images, $\mathbf{P}_i,\mathbf{P}_j$ their camera poses, and $\tau$ a generic triangulation method. Usually, $\Tilde{\mathbf{x}}_i, \Tilde{\mathbf{x}}_j$ pairs are identified in advance by extracting features and matching them. Eventually, global optimization is carried out with bundle adjustment, minimizing the reprojection error $E$ 

\small\begin{equation}
    E = \sum_j || \pi(\mathbf{P}_i,\mathbf{X}_i) - \mathbf{x}_j ||_2^2
\end{equation}\normalsize
with $\pi$ being the projection function from 3D to image space. SfM algorithms -- COLMAP \cite{schoenberger2016sfm} in particular -- are the foundation for bootstrapping GS optimization, providing both the camera poses and the 3D points for initializing the Gaussians. 

\textbf{Depth Completion (DC).} This method aims to recover a dense depth map from a set of sparse measurements $\mathbf{X}_i$, usually guided by a color image $\mathbf{I}_i$, with a network $\Theta_\text{DC}$: 

\small\begin{equation}
    \mathbf{D}_\text{DC}(\mathbf{I}_i) = \Theta_\text{DC}(\mathbf{I}_i, \mathbf{X}_i)
\end{equation}\normalsize
In our setting, a DC model can process the sparse set of points estimated by COLMAP and projected over the images, similarly to \cite{roessle2022dense} in principle. From a practical perspective, the availability of a DC model capable of generalizing across scenes and levels of sparsity is crucial for this purpose -- although ignored in \cite{roessle2022dense}. 

\textbf{Multi-View Stereo (MVS).} A dense depth map can be obtained by matching pixels across multiple, posed images along epipolar lines. 
This task is nowadays tackled with deep networks as well \cite{yao2018mvsnet}, with a generic model $\Theta_\text{MVS}$ processing a reference image $\mathbf{I}_i$ and a set of $N$ source views, given their poses

\small\begin{equation}
    \mathbf{D}_\text{MVS}(\mathbf{I}_i) = \Theta_\text{MVS}(\mathbf{I}_i,\mathbf{P}_i, \{ \mathbf{I}_j,\mathbf{P}_j \}_{j}^{N} )
\end{equation}\normalsize
Despite the outstanding accuracy reached in the last years, MVS networks still struggle when dealing with sparse views with limited overlap, and may suffer from generalization issues.

\begin{figure*}[t]
    \centering
    \renewcommand{\tabcolsep}{1pt}
    \begin{tabular}{ccccc}
        \scriptsize{Rendered Image} & \scriptsize{Rendered Depth} & \scriptsize{Rendered Right} & \scriptsize{Stereo Depth} & \scriptsize{GT Depth}\\
        \includegraphics[width=0.18\textwidth]{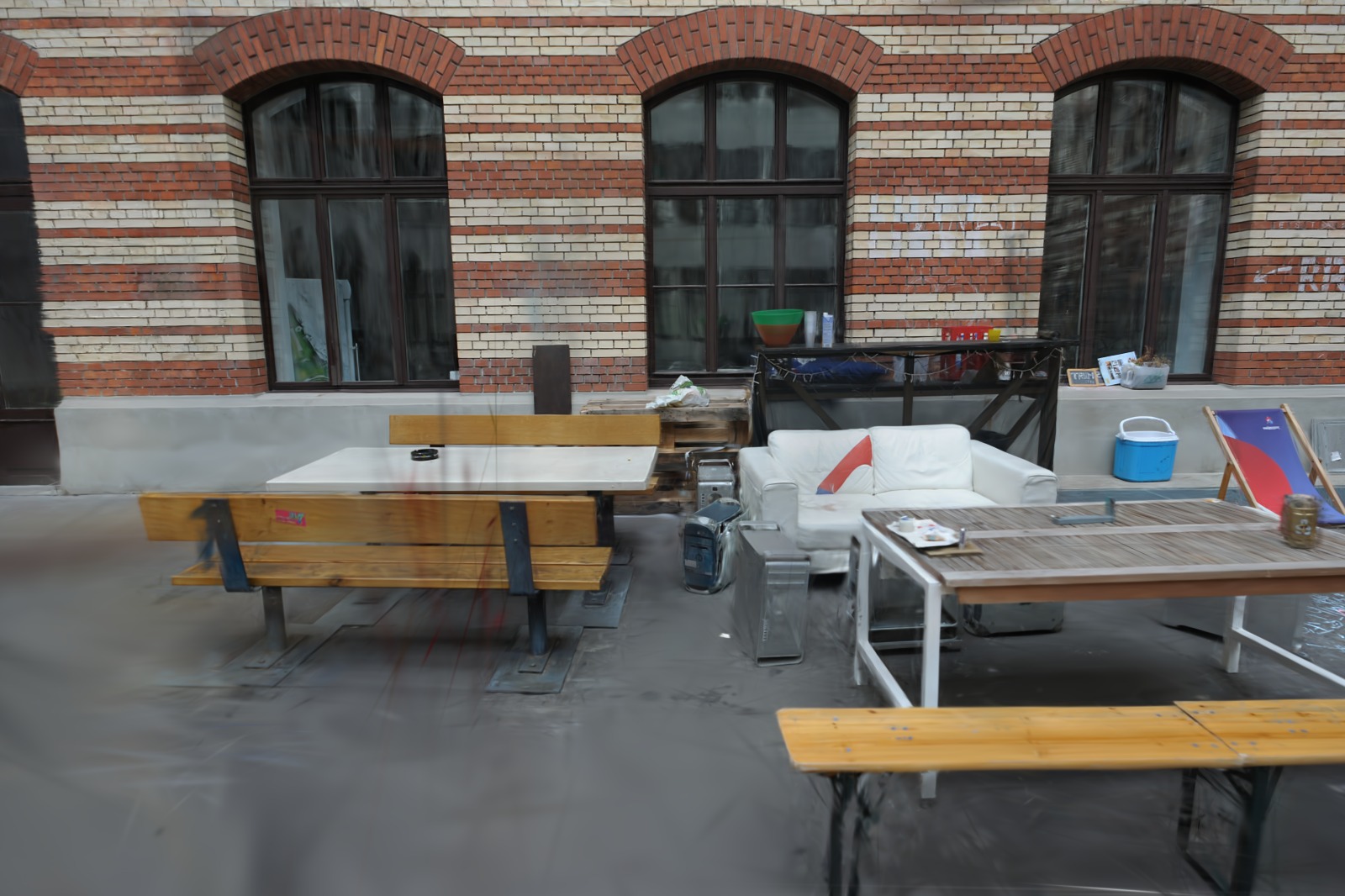} & 
        \includegraphics[width=0.18\textwidth]{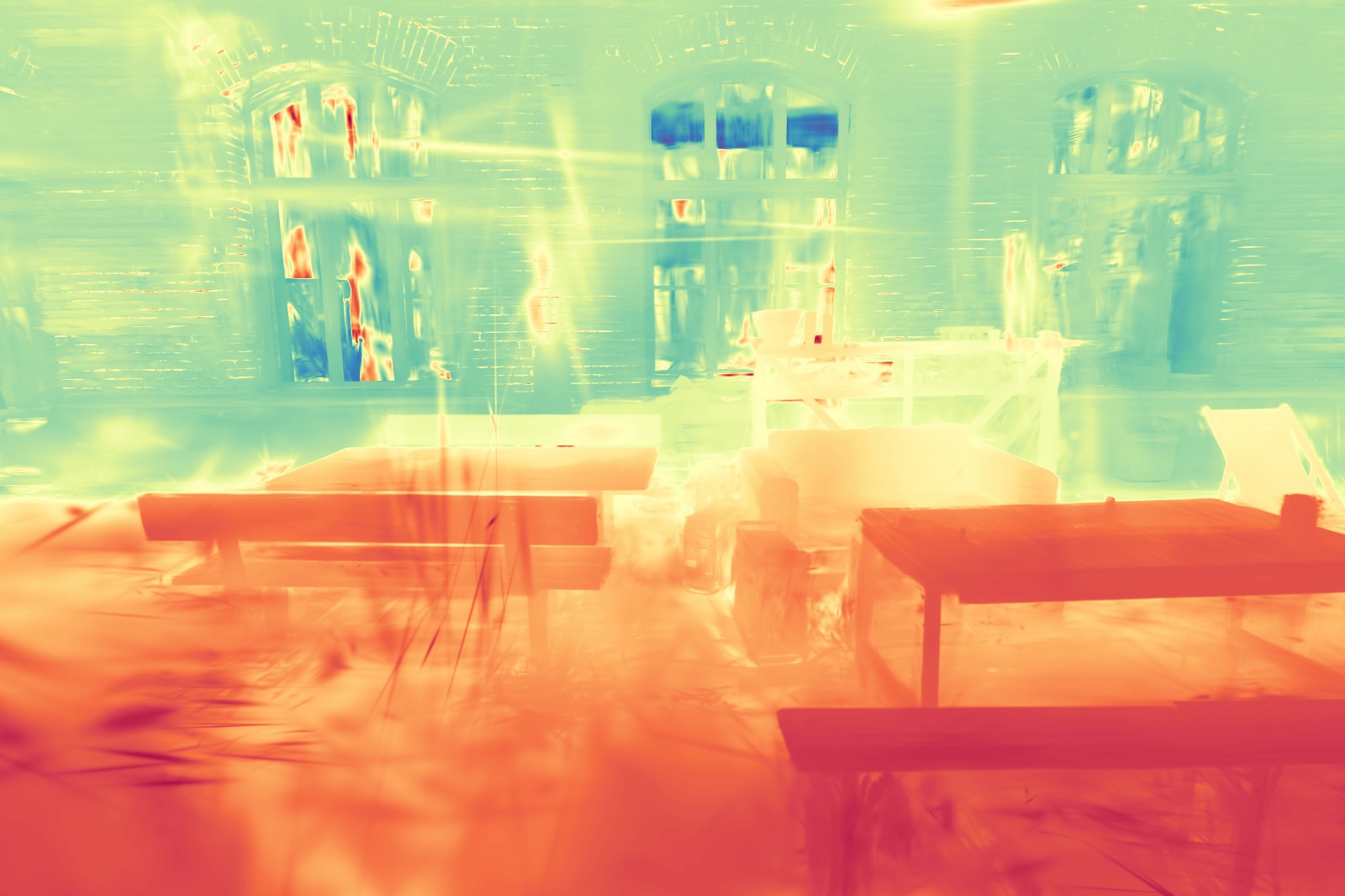} &
        \includegraphics[width=0.18\textwidth]{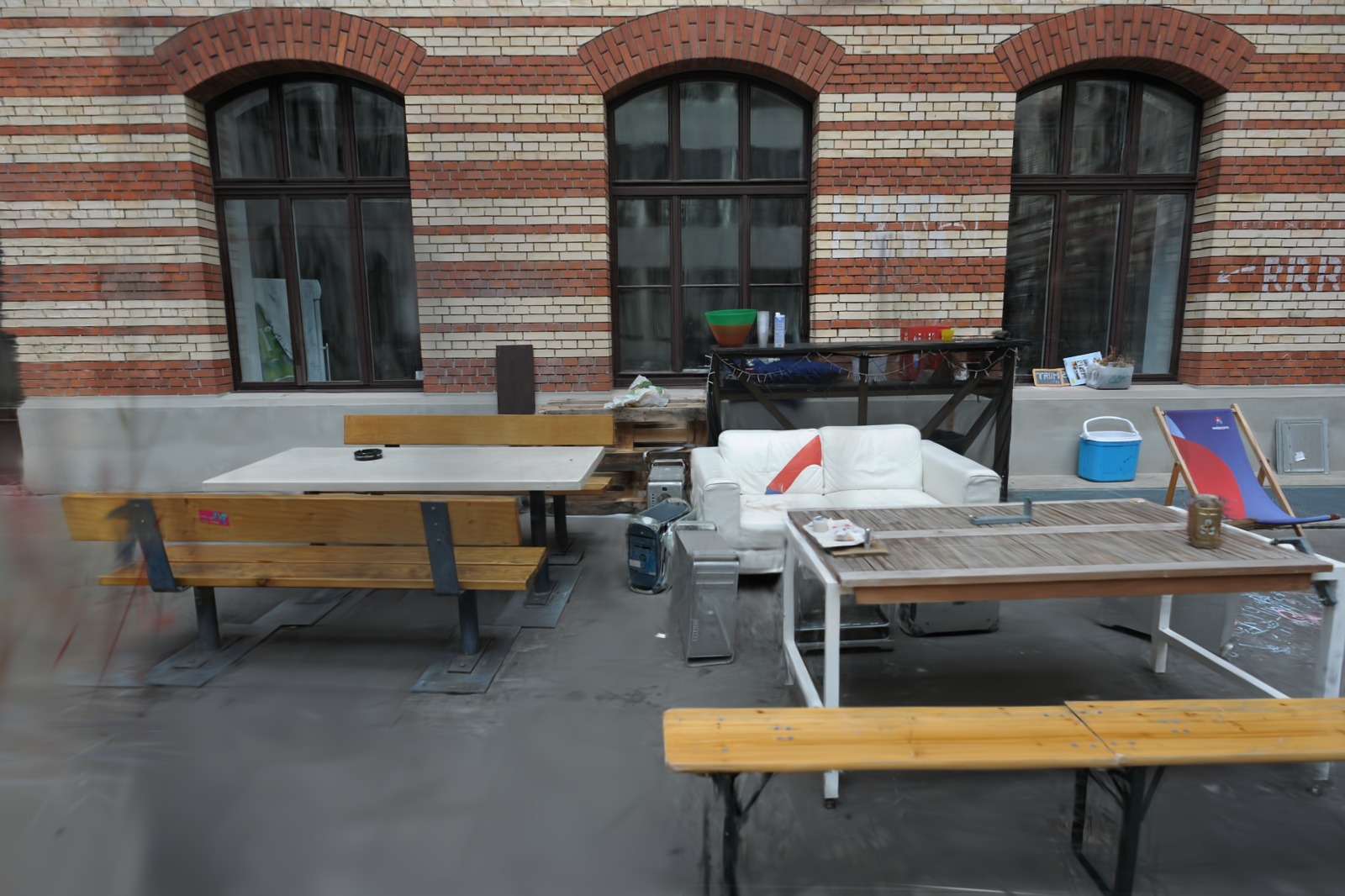} &
        \includegraphics[width=0.18\textwidth]{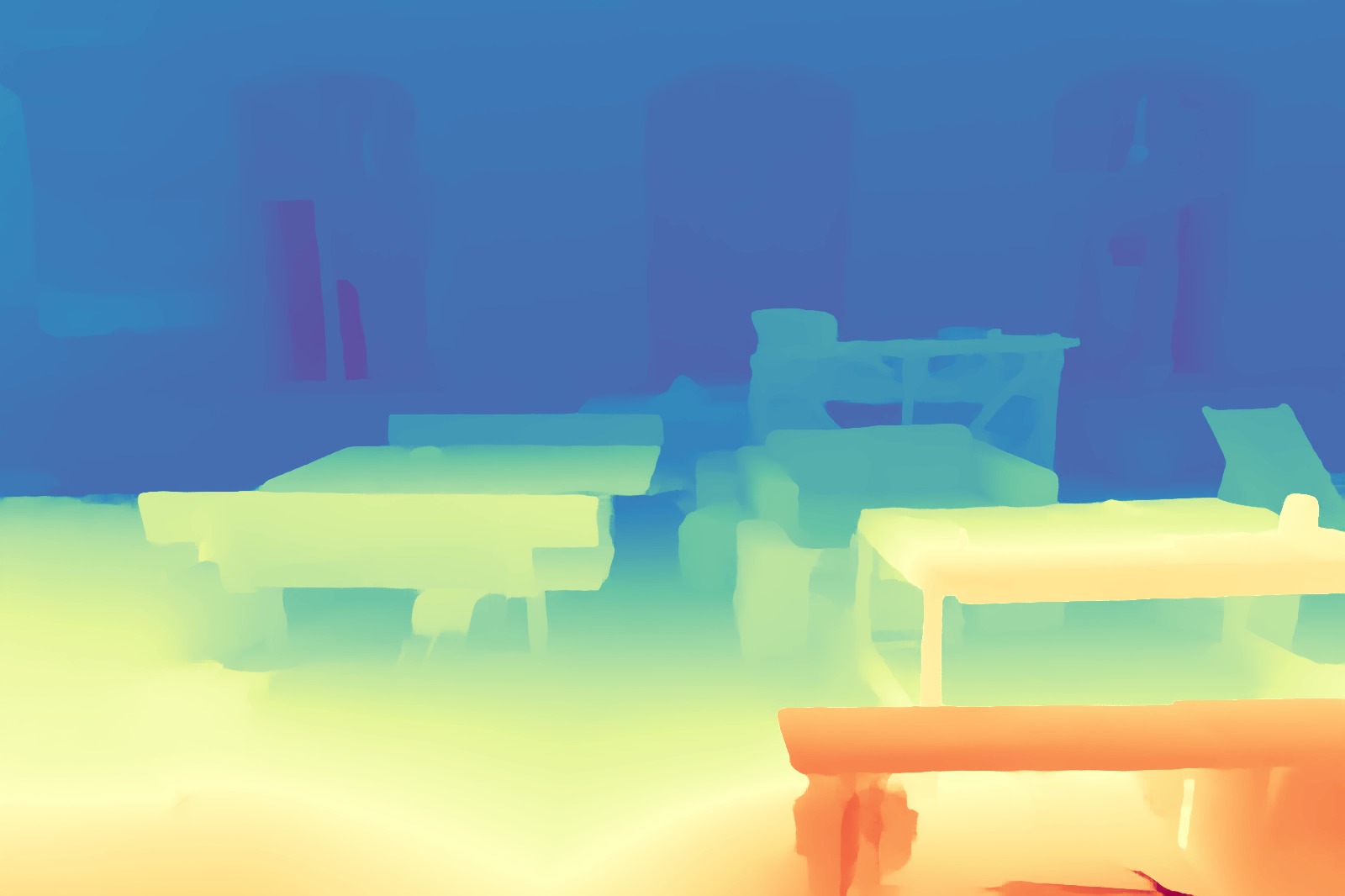} &
        \includegraphics[width=0.18\textwidth]{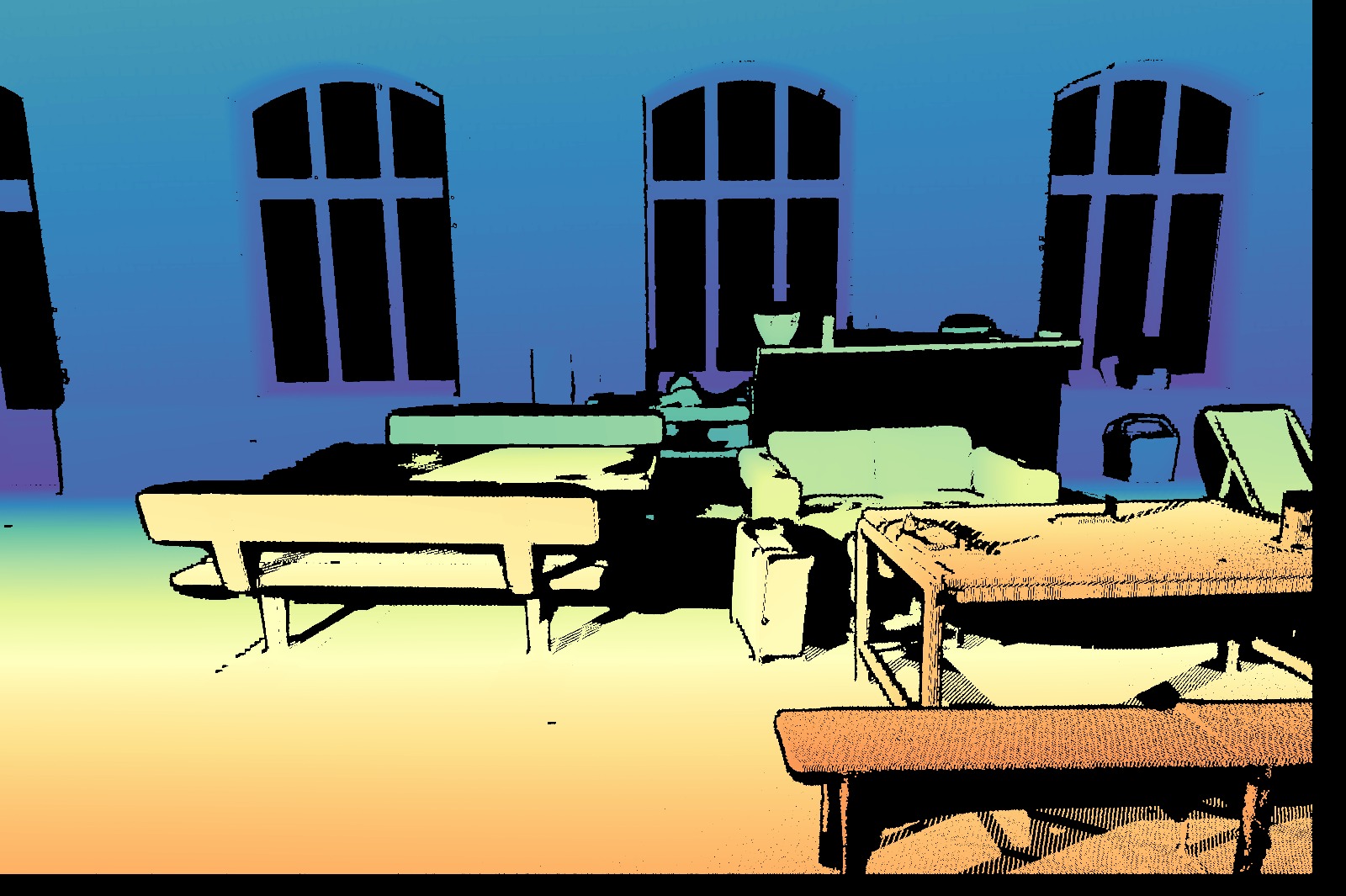} \\
        \includegraphics[width=0.18\textwidth]{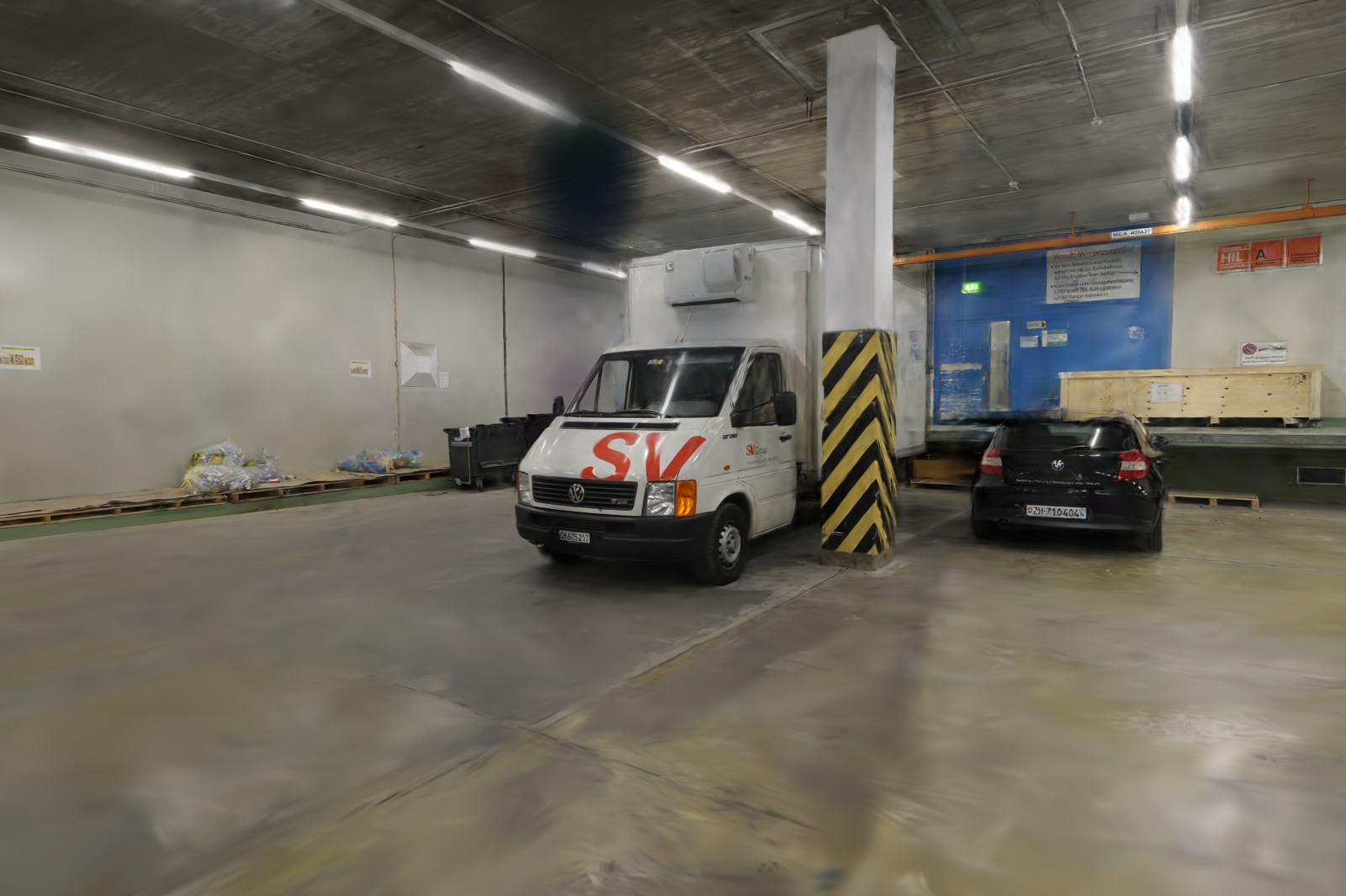} & 
        \includegraphics[width=0.18\textwidth]{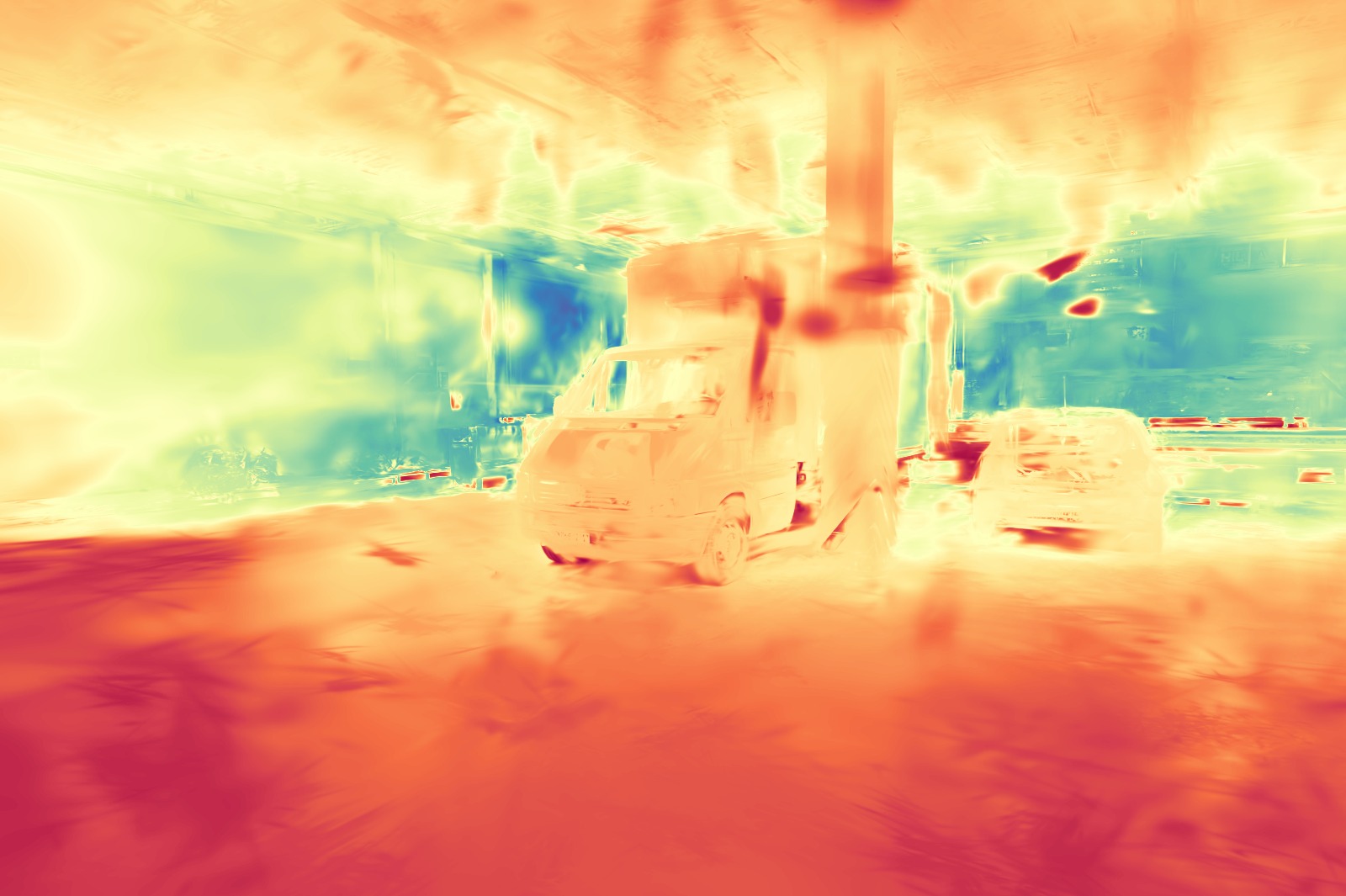} &
        \includegraphics[width=0.18\textwidth]{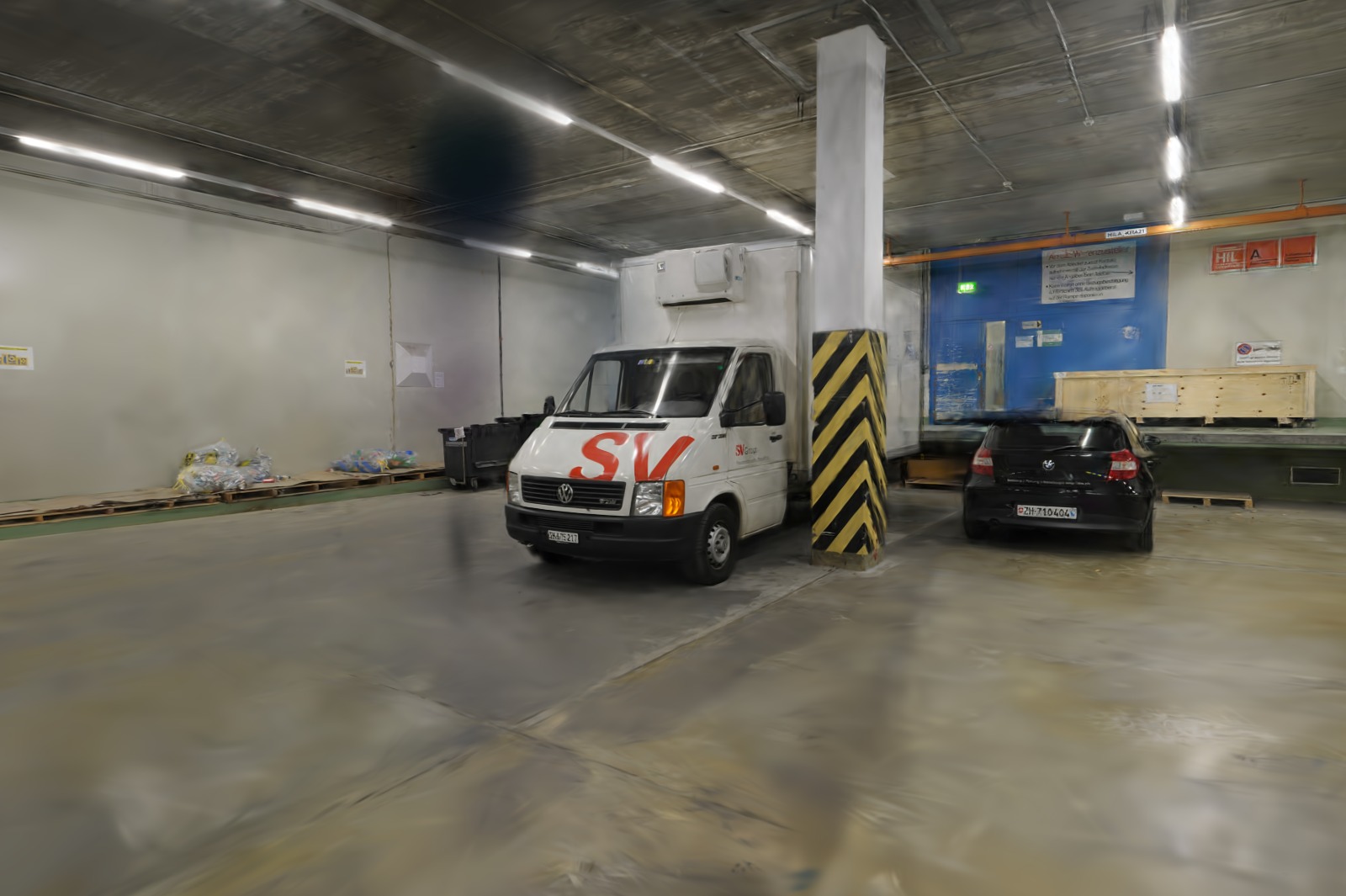} &
        \includegraphics[width=0.18\textwidth]{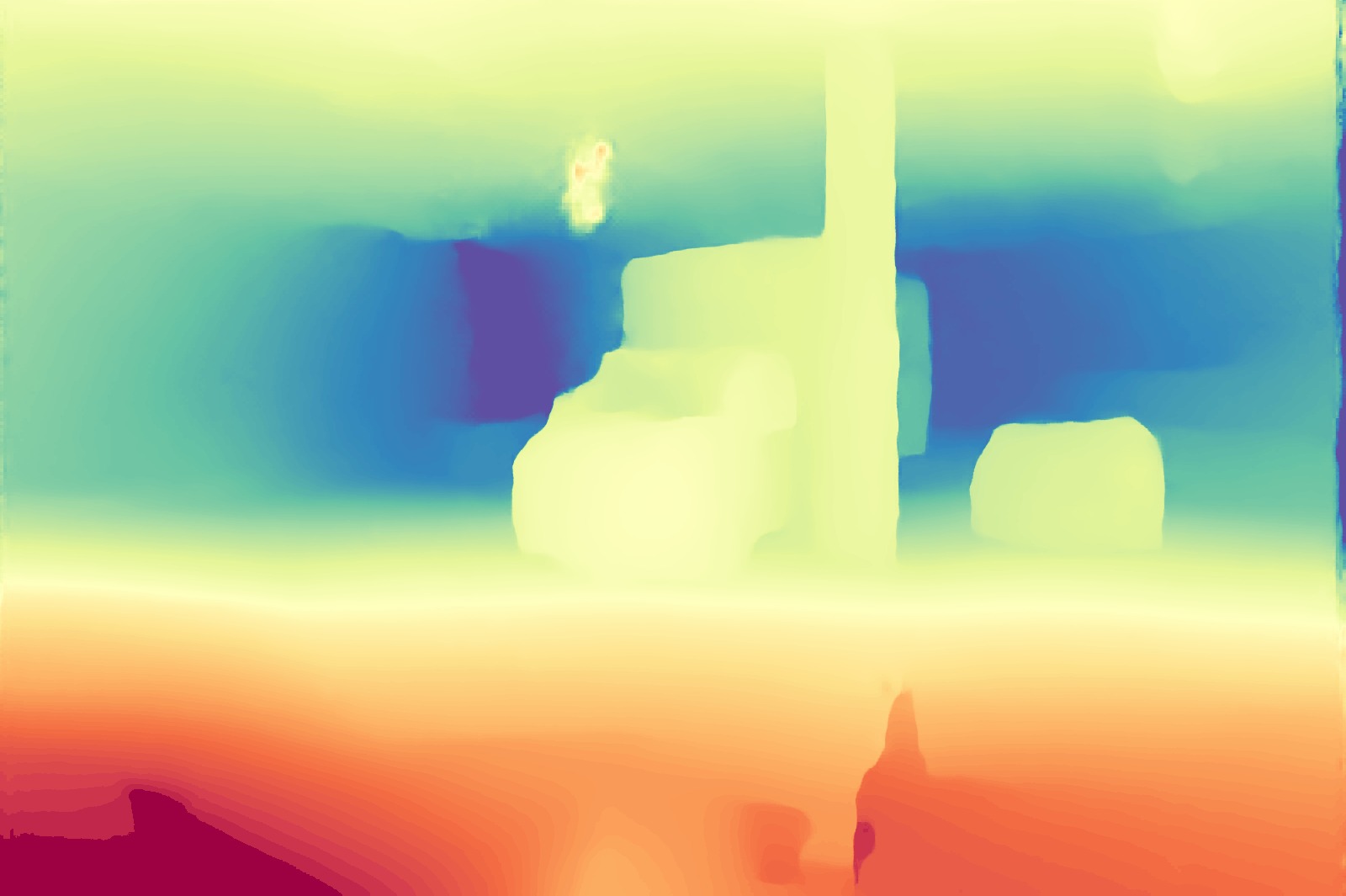} &
        \includegraphics[width=0.18\textwidth]{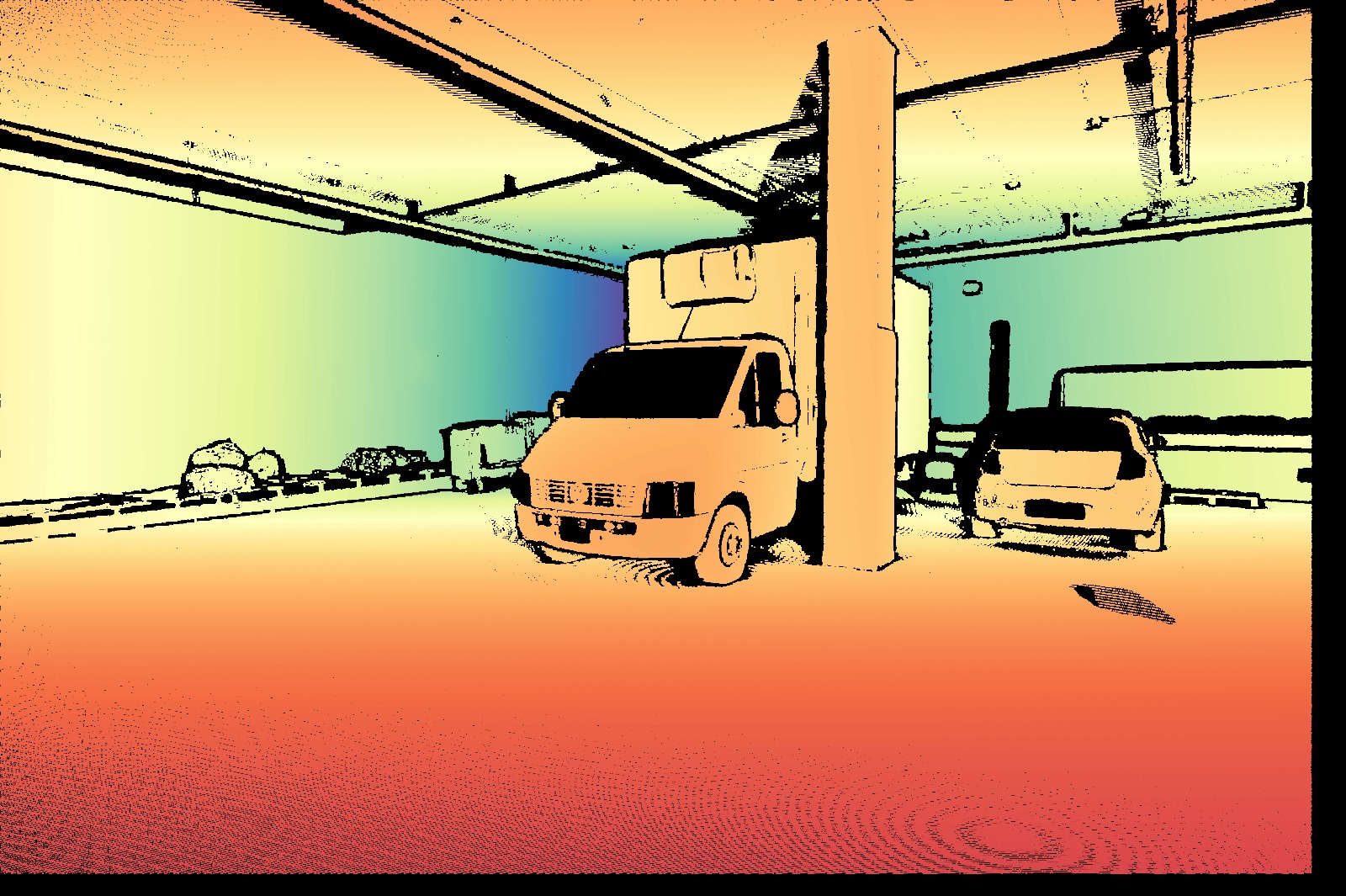} \\
    \end{tabular}
    \caption{\textbf{Depth priors retrieved from stereo.}  
    Vanilla GS produces noisy and inaccurate depths, yet can render stereo pairs for getting strong depth priors as additional supervision.}
    \label{fig:stereo-depth}
\end{figure*}

\textbf{Monocular Depth Estimation (MDE).} With the rise of deep learning, estimating the depth of a single image has become a reality. 
Nowadays, state-of-the-art models are trained over millions of images to predict \textit{affine-invariant} depth \cite{Ranftl2021,Ranftl2022}:

\small\begin{equation}
    \mathbf{D}_\text{MDE}(\mathbf{I}_i) = m\cdot \Theta_\text{MDE}(\mathbf{I}_i) + q
\end{equation}\normalsize
where $m$ and $q$ are respectively \textit{scale} and \textit{shift} factors required to recover the effective scale within the relative depth map predicted by the model $\Theta_\text{MDE}$. In our specific setting, $m$ and $q$ can be directly derived by fitting the predicted depth map on the COLMAP depth points through least squares. A concurrent work \cite{chung2023depth} also follows this strategy to regularize GS optimization -- yet without measuring its impact on the underlying geometry.
\section{Self-Evolving Depth-Supervised GS from Stereo}
\label{sec:stereo}

In this section, we introduce an alternative strategy to obtain dense depth priors and improve the optimization process of GS. 
We begin with the empirical observation that, despite its inaccurate underlying geometry, GS can render images that exhibit \textit{geometric consistency}. 

This means that a trained GS can render frames over which we can run conventional depth-from-images algorithms to retrieve quite accurate depth priors for supervising the GS itself.
Purposely, the simplest strategy consists of rendering \textit{rectified stereo pairs} -- i.e., images captured from two viewpoints shifted by a horizontal offset -- and then estimating depth through triangulation from disparity.
\cref{fig:stereo-depth} shows qualitative evidence of the effectiveness of this strategy: while vanilla GS renders noisy depth maps, a pre-trained stereo model can generate much better depth maps from stereo pairs rendered by vanilla GS itself.

For this purpose, given any camera pose $\mathbf{P}_i$, we can derive a corresponding right viewpoint with pose $\mathbf{R}_i$ in a fictitious stereo configuration, according to an arbitrary baseline $b$: 
\begin{equation}
    \mathbf{R}_i = 
    \begin{pmatrix}
        \mathbb{I} & \mathbf{t} \\
        0 & 1 \\
    \end{pmatrix} \cdot \mathbf{P}_i \quad\quad \text{with} \quad \textbf{t} = 
    \begin{pmatrix} b & 0 & 0 \end{pmatrix}^{\top}
\end{equation}
Then, for each image $\mathbf{I}_i$ in the training set, we can render a corresponding right frame $\mathbf{I}^r_i$, estimate disparity with a stereo network $\Theta$ and use the focal length $f$ to triangulate depth:

\begin{equation}
    \mathbf{D}_\text{stereo}(\mathbf{I}_i) = \frac{f \cdot b}{ {\Theta}(\mathbf{I}_i,\mathbf{I}^r_i) }
\end{equation}

\begin{figure*}[t]
    \centering
    \includegraphics[ width=\linewidth]{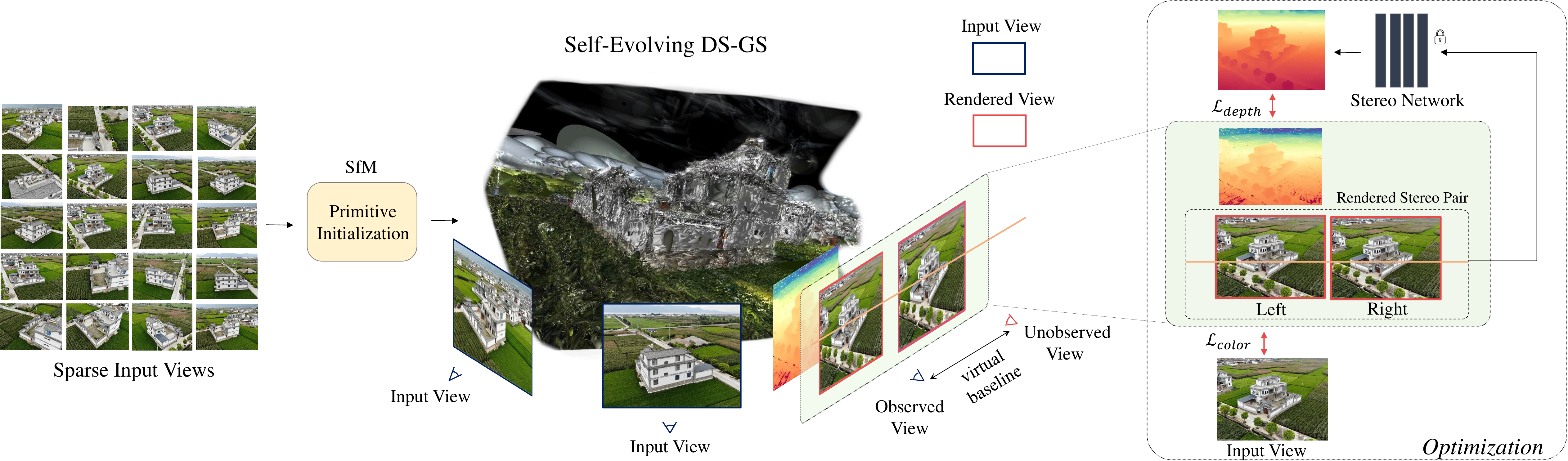}
    \vspace{-0.3cm}
    \caption{\textbf{Self-Evolving DS-GS pipeline.} As soon as GS can render stable images, we render \textit{stereo pairs}, estimate depth with a pre-trained network, and use it to compute $\mathcal{L}_{depth}$.
    }
    \label{fig:framework}
\end{figure*}

\cref{fig:framework} provides an overview of our approach. During GS training, we can start exploiting this strategy only after the model can render good-quality images already -- i.e., after $T$ steps. 
Furthermore, 
as disparity estimation requires a non-negligible extra computation, 
we cache disparity maps as soon as they are computed the first time and reuse them in the subsequent steps; as long as the quality of rendered images increases with training, we set a refresh interval $\tau$ for rendering again the stereo pairs and updating the disparity priors.

\textbf{Why stereo?} 
Our strategy would work to render views for MVS networks as well, however:
i) as we aim at minimizing the overhead during GS optimization, frame pairs are the minimum amount of information required to derive depth from images through geometry; ii) tuning a single parameter for rendering the novel, arbitrary views -- i.e., the horizontal baseline $b$ -- is simpler than tuning 6-DoF poses; iii) state-of-the-art stereo networks excel at domain generalization, whereas we observed this is often not true for MVS networks.

\section{Experiments}
\label{sec:experiments}

In this section, we present the approaches we selected to obtain depth priors and the datasets used in our experiments. Next, we describe the implementation details of our framework. Finally, we report our findings.

\subsection{Depth Priors Settings}

To evaluate the effectiveness of the different strategies for extracting depth priors from images, we select one representative method for each. \textbf{SfM:} We use COLMAP \cite{schoenberger2016sfm}, as it is already used for computing camera poses and the initial 3D points from which GS optimization is bootstrapped. \textbf{DC:} We select VPPDC \cite{bartolomei2023revisiting} since, to the best of our knowledge, it is the only depth completion approach proposed for cross-domain generalization. \textbf{MDE:} We use ZoeDepth \cite{bhat2023zoedepth}, with the weights provided by the authors. The same model has been used in a concurrent work \cite{chung2023depth}, allowing us to asses its effectiveness against alternative depth priors sources.
\textbf{MVS:} We use CER-MVS \cite{ma2022multiview}, as it shows promising generalization. We use BlendedMVS weights for tests on ETH3D/ScanNet++, and DTU weights on BlendedMVS, to avoid overlap with training data. 
\textbf{Stereo (Self-Evolving):} We use the RAFT-Stereo \cite{lipson2021raft} variant trained for the Robust Vision Challenge -- iRAFT-Stereo\_RVC \cite{jiang2022iraft}.

\subsection{Datasets}
 
We select three datasets providing ground-truth depth, instrumental for our studies. We will appreciate how the differences in the three will impact the results by different methods.

\textbf{ETH3D.} It is a real-world dataset, providing images and ground truth depth at about 24 Megapixels. We use all of the 13 training scenes of the high-resolution set, having 14 to 76 images. We use the provided undistorted images, camera poses, and point clouds. 
Following vanilla GS settings, images are resized to have 1600 width before training. We manually split each scene into training and test sets (please check the \textbf{supplementary} for details of the splits). We align (distorted) ground truth depth to undistorted images for evaluation.

\textbf{ScanNet++.} It is a real-world dataset with high-fidelity 3D geometry and high-resolution RGB images of
indoor scenes. We perform our experiments on 2 of the randomly selected scenes, due to the large number of sequences. We undistort the fisheye images and depth maps using the provided official toolkit. The scenes contain 291 and 399 images at a resolution of 1752$\times$1168, and we only use every 10th image for training and the rest for testing.

\textbf{BlendedMVS.} It is a semi-synthetic dataset. Due to the large number of sequences, we randomly select 4 of them and perform experiments on these sequences.  Since this dataset does not provide the point clouds, we run COLMAP on the images using the given camera poses to obtain the point clouds for bootstrapping GS. We use the images at their original resolution of 768$\times$576, 
counting 75 to 212 frames per scene, with every 4th used for testing.

\subsection{Implementation Details}

We implement our self-evolving GS starting from \cite{kerbl20233d}. 
The loss for DS-GS is defined as:
\begin{equation}
    \mathcal{L} = (1-\lambda_1)||\mathbf{I} - \hat{\mathbf{I}}||_1 + \lambda_1 \text{D-SSIM}(\mathbf{I}, \hat{\mathbf{I}}) + \lambda_2 ||\mathbf{D_k(\mathbf{I})}  - \mathbf{\hat{D}}||_1
\end{equation}
where $\mathbf{I}$ is the original image, $\mathbf{\hat{I}}$ and $\mathbf{\hat{D}}$ are the rendered image and depth maps, respectively, and $\mathbf{{D}_k}$ is the depth map, obtained through one of the proposed methods. e.g., for our self-evolving method $\mathbf{{D}_k}= \mathbf{D}_\text{Stereo}$.  
We set $\lambda_1 = 0.2$ and set $\lambda_2=0.01$ for the BlendedMVS dataset and $\lambda_2=0.1$ for the ETH3D and ScanNet++ datasets.
Note that for our self-evolving method, we set $\lambda_2=0$ for all iterations before the starting step $T$. On the BlendedMVS dataset, we use $T=17K$ and train the models for $20K$ iterations, while for the ETH3D and ScanNet++ datasets, we set $T=7K$ and train for $11K$ iterations. In all datasets, we set the refresh interval to $\tau=100$ and randomly sample $b$ from an interval  
(see \textbf{supplementary material}). 
We perform all of our experiments on a single V100 GPU.
\subsection{Results}
In this section, we report the outcome of our experiments. 
In each table, we highlight the \colorbox{First}{\textbf{first}}, \colorbox{Second}{\textbf{second}} and \colorbox{Third}{\textbf{third}} -best results.

\begin{table}[t]
\centering
\begin{adjustbox}{max width=\textwidth}
    \renewcommand{\tabcolsep}{10pt}
    \begin{tabular}{cl ccc ccc}
        & & \multicolumn{3}{c}{Depth} & \multicolumn{3}{c}{View Synthesis} \\
        \cmidrule(lr){1-2} \cmidrule(lr){3-5} \cmidrule(lr){6-8}
        & Method & Abs. Rel. $\downarrow$& RMSE  $\downarrow$& $\delta < 1.25$  $\uparrow$ & SSIM  $\uparrow$& PSNR  $\uparrow$& LPIPS  $\downarrow$ \\
        \cmidrule(lr){1-2} \cmidrule(lr){3-5} \cmidrule(lr){6-8}
        & GS \cite{kerbl20233d} & 0.211 & 1.698 & 0.652 & 0.7425 & 20.4029 & 0.3385  \\

        \addlinespace
        \hdashline
        \addlinespace
        \multirow{5}{*}{\rotatebox[origin=l]{90}{DS-GS}} & + SfM\cite{schoenberger2016sfm} & \trd 0.109 & \snd 0.870 & \trd 0.844 & \trd 0.7561 & \trd 21.7224 &  0.3261  \\
        & + DC\cite{bartolomei2023revisiting} & 0.148 & 1.272 & 0.828 & 0.7557 & 21.5206 & \trd 0.3248  \\
        & + MDE\cite{bhat2023zoedepth} & 0.153 & 1.204 & 0.793 & 0.7475 & 21.3104 & 0.3377  \\        
        & + MVS\cite{ma2022multiview} & \snd 0.094 & \trd 1.031 & \snd 0.914 & \snd 0.7692 & \snd 22.2806 & \fst 0.3105  \\
        & + \textbf{Self-Evolving (ours)} & \fst 0.057 & \fst 0.599 & \fst 0.942 & \fst 0.7704 & \fst 22.2825 & \snd 0.3141  \\

        \addlinespace
        \hdashline
        \addlinespace
        & + Oracle (GT depth) & 0.020 & 0.317 & 0.980 & 0.7764 & 22.4669 & 0.3009  \\
        \hline
    \end{tabular}
    \end{adjustbox}
    \caption{\textbf{Quantitative Results on ETH3D.}}
    \label{tab:eth3d}
\end{table}

\begin{table}[t]
    \centering
\begin{adjustbox}{max width=\textwidth}
    \renewcommand{\tabcolsep}{10pt}
    \begin{tabular}{cl ccc ccc}
        & & \multicolumn{3}{c}{Depth} & \multicolumn{3}{c}{View Synthesis} \\
        \cmidrule(lr){1-2} \cmidrule(lr){3-5} \cmidrule(lr){6-8}
        & Method & Abs. Rel. $\downarrow$& RMSE  $\downarrow$& $\delta < 1.25$  $\uparrow$ & SSIM  $\uparrow$& PSNR  $\uparrow$& LPIPS  $\downarrow$ \\
        \cmidrule(lr){1-2} \cmidrule(lr){3-5} \cmidrule(lr){6-8}
        & GS \cite{kerbl20233d} & 0.154 & 0.416 & 0.735 & 0.9162 & 27.7907 & 0.1587  \\

        \addlinespace
        \hdashline
        \addlinespace
        \multirow{5}{*}{\rotatebox[origin=l]{90}{DS-GS}} & + SfM\cite{schoenberger2016sfm} & \trd 0.104 & \trd 0.295 & \trd 0.860 &  0.9131 & \trd 27.9140 &  0.1663  \\
        & + DC\cite{bartolomei2023revisiting}  & 0.144 & 0.381 & 0.813 & \trd 0.9145 & 27.5081 & \trd 0.1659 \\
        & + MDE\cite{bhat2023zoedepth} & \snd 0.083 & \snd 0.242 & \snd 0.926 & \fst 0.9168 & \snd 28.0568 & \fst 0.1588  \\        
        & + MVS\cite{ma2022multiview} &  0.152 & 0.437 & 0.824 &  0.9138 & 27.3536 & 0.1699  \\
        & + \textbf{Self-Evolving (ours)} & \fst 0.068 & \fst 0.222 & \fst 0.928 & \snd 0.9165 & \fst 28.1488 & \snd 0.1601  \\

        \addlinespace
        \hdashline
        \addlinespace
        & + Oracle (GT depth) & 0.024 & 0.103 & 0.983 & 0.9199 & 28.6413 & 0.1539  \\
        \hline
    \end{tabular}
    \end{adjustbox}
    \caption{\textbf{Quantitative Results on ScanNet++.}}
    \label{tab:scannet}
\end{table}

\textbf{ETH3D.} \cref{tab:eth3d} shows quantitative results on ETH3D. 
We can appreciate how any DS-GS variant improves over vanilla GS, both in terms of depth and color rendering quality.
Using COLMAP yields the third-best result and unveils an interesting finding: vanilla GS optimization is sub-optimal since the very same 3D points used to bootstrap Gaussians can provide additional supervision for free. 
Nonetheless, DC fails at improving over COLMAP, because of the very few SfM points being insufficient for obtaining a good-quality dense depth map -- see the \textbf{supplementary material} for qualitative examples. 
MVS ranks second both in terms of depth and color quality, while our self-evolving DS-GS largely outperforms it in terms of depth estimation, slightly improving color quality in SSIM and PSNR as well.

\textbf{ScanNet++.} \cref{tab:scannet} collects results on ScanNet++. At first glance, we can confirm the superiority of our approach in terms of depth metrics, while resulting almost equivalent to MDE on color metrics.
Interestingly, MDE significantly outperforms SfM and other methods in this setting. This is caused by the lack of texture in these scenes, on which COLMAP extracts few 3D points (see \textbf{supplementary material}) and thus provides poor supervision.

\textbf{BlendedMVS.} 
\cref{tab:blendedmvs} collects results on BlendedMVS. 
On these semi-synthetic images, SfM can extract very dense matches, resulting in much stronger supervision as well as a much simpler completion task for the DC model (see \textbf{supplementary material}). Indeed, DC ranks second in both depth and color metrics, with SfM ranking first in rendering quality.
On the contrary, our solution is, again, the absolute winner in terms of depth accuracy, while MDE and MVS fail at improving the baseline: we ascribe this to generalization issues.

\textbf{Qualitative Results.} We conclude this section by reporting some qualitative comparisons. 
\cref{fig:qual_eth3d} collects three samples from ETH3D dataset. At the very top, we show images and depth maps obtained by the vanilla GS, with several artifacts appearing in any of the three examples, followed by results yielded by using SfM or our strategy. 
Conversely to SfM, our self-evolving GS can consistently improve both rendered images and depth maps. 
Finally, at the very bottom, we report ground-truth images and depth maps as a reference. We report more qualitative results in the \textbf{supplementary material.}

\begin{table}[t!]
    \centering
    \renewcommand{\tabcolsep}{10pt}
    \begin{adjustbox}{max width=\textwidth}
    \begin{tabular}{cl ccc ccc}
        & & \multicolumn{3}{c}{Depth} & \multicolumn{3}{c}{View Synthesis} \\
        \cmidrule(lr){1-2} \cmidrule(lr){3-5} \cmidrule(lr){6-8}
        & Method & Abs. Rel. $\downarrow$ & RMSE  $\downarrow$& $\delta < 1.25$  $\uparrow$ & SSIM  $\uparrow$& PSNR  $\uparrow$& LPIPS  $\downarrow$ \\
        \cmidrule(lr){1-2} \cmidrule(lr){3-5} \cmidrule(lr){6-8}
        & GS \cite{kerbl20233d}  & 0.058 & 7.041 & 0.933 & 0.6301 & 21.6160 & 0.2729  \\
        \addlinespace
        \hdashline
        \addlinespace
        \multirow{5}{*}{\rotatebox[origin=l]{90}{DS-GS}} & + SfM\cite{schoenberger2016sfm} & \snd 0.021 & \trd 3.910 & \trd 0.990 & \fst 0.6389 & \fst \fst 22.1409 & \fst 0.2644  \\
        & + DC\cite{bartolomei2023revisiting} & \snd 0.021 & \snd 3.719 & \snd 0.991 & \snd 0.6378 & \snd 21.9899 & \snd 0.2648  \\
        & + MDE\cite{bhat2023zoedepth} & 0.113 & 12.141 & 0.840 & 0.6142 & 21.1857 & 0.2867  \\
        & + MVS\cite{ma2022multiview} & 0.065 & 10.316 & 0.914 & 0.6021 & 20.8971 & 0.2944  \\ 
        & + \textbf{Self-Evolving (ours)} & \fst 0.020 & \fst 3.714 & \fst 0.992 & \trd 0.6377 & \trd 21.9734 & \trd 0.2696  \\
        \addlinespace
        \hdashline
        \addlinespace
        & + Oracle (GT depth) & 0.013  & 2.645 & 0.993 & 0.6480 & 22.2282 & 0.2575  \\
        \hline
    \end{tabular}
        \end{adjustbox}
        \caption{\textbf{Quantitative Results on BlendedMVS.}}
    \label{tab:blendedmvs}

\end{table}

\begin{figure*}[t!]
    \centering
    \renewcommand{\tabcolsep}{1pt}
    \begin{tabular}{rcccccccc}    
    \rotatebox[origin=l]{90}{\scriptsize{\quad GS}} & 
        \includegraphics[width=0.15\textwidth]{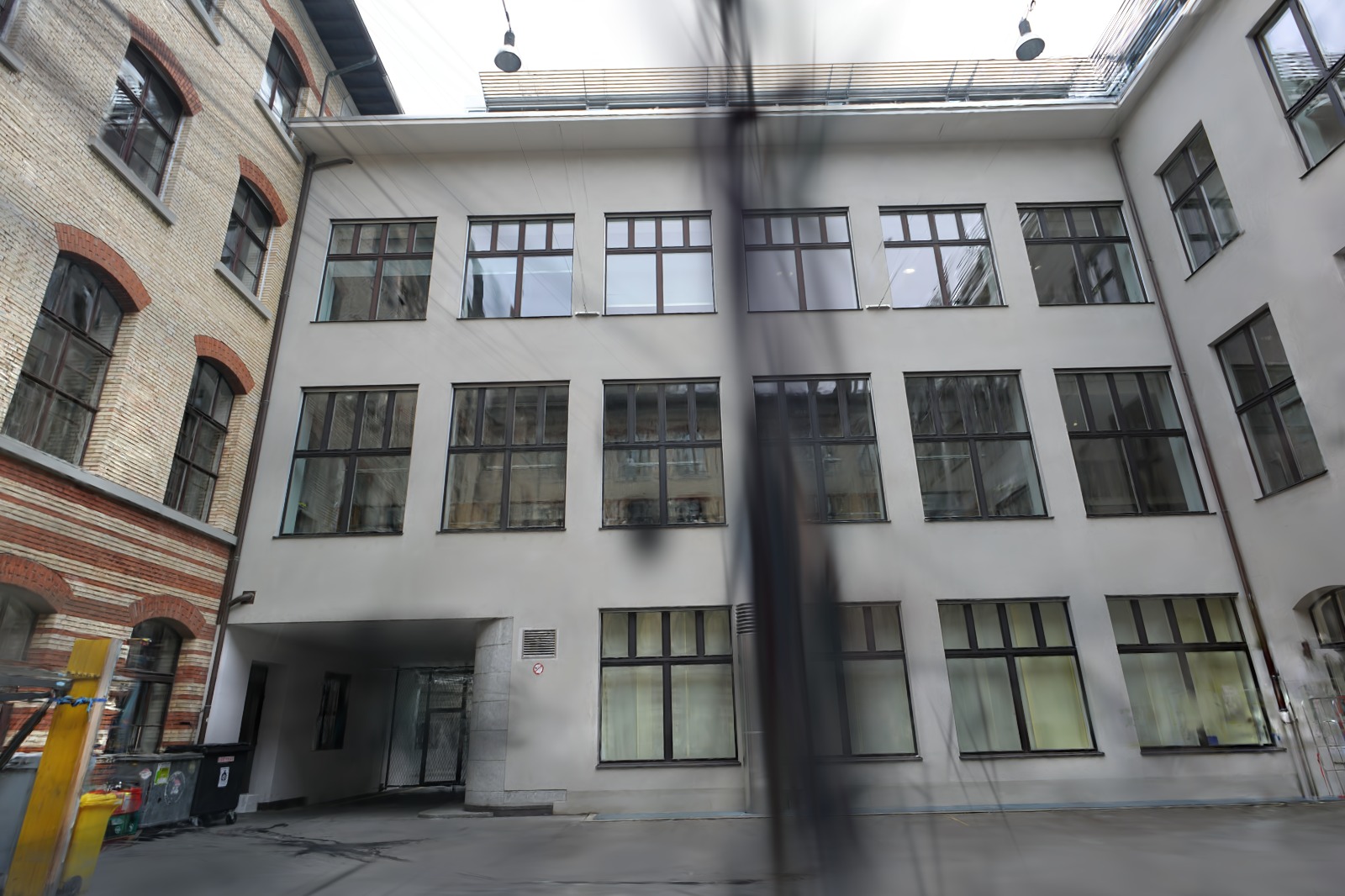} & 
        \includegraphics[width=0.15\textwidth]{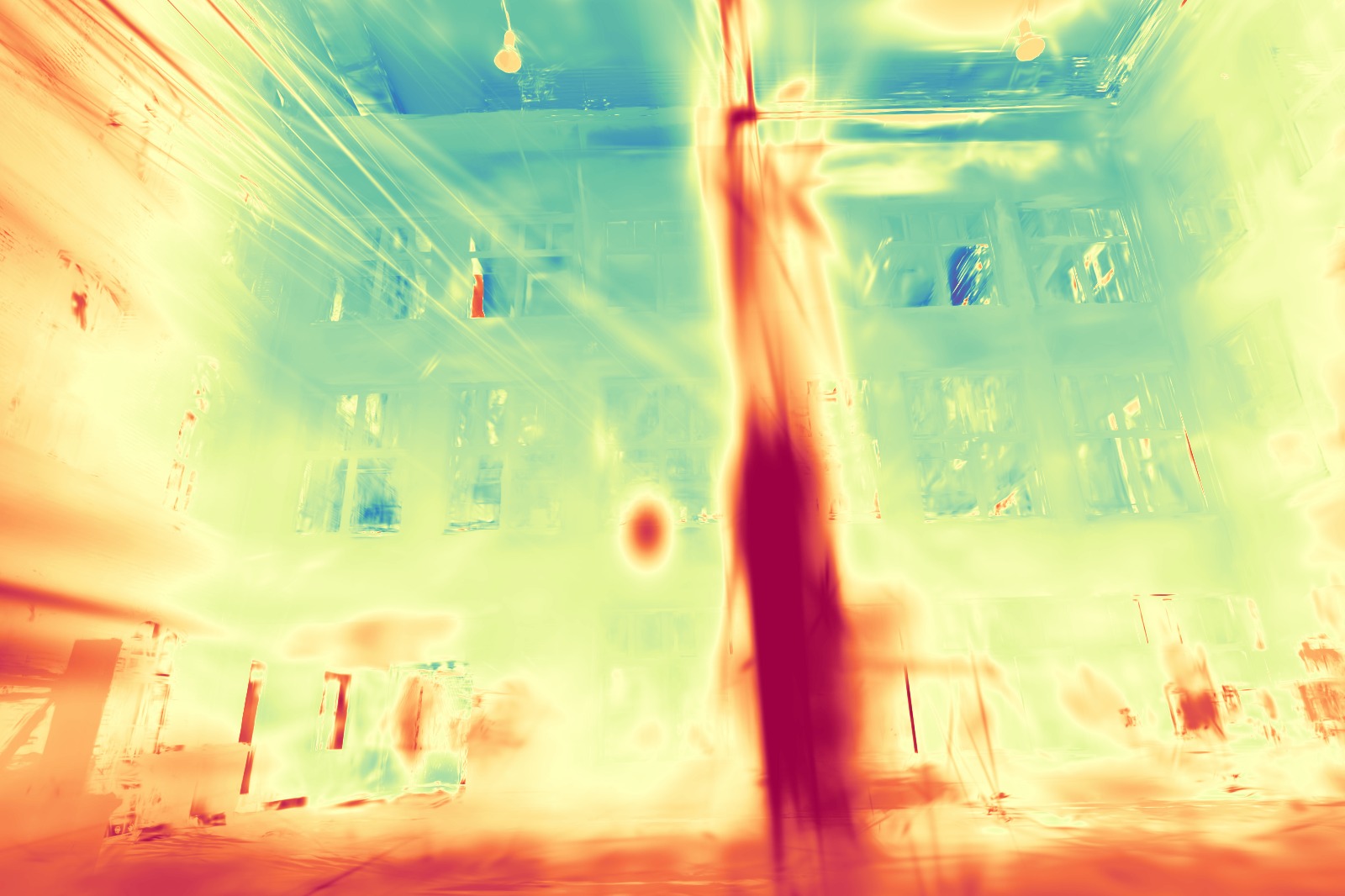} & \hspace{0.3cm} &
        \includegraphics[width=0.15\textwidth]{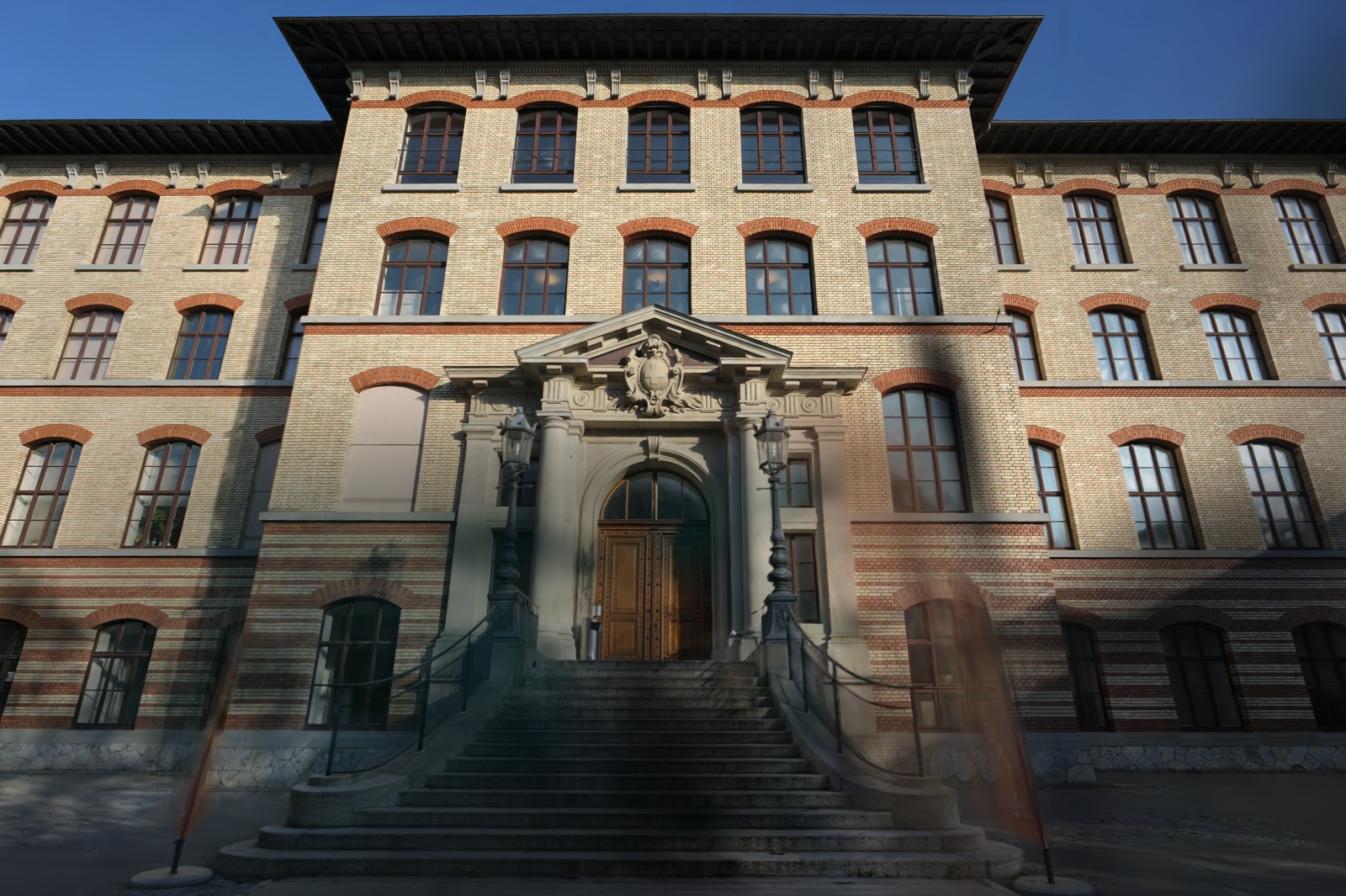} &
        \includegraphics[width=0.15\textwidth]{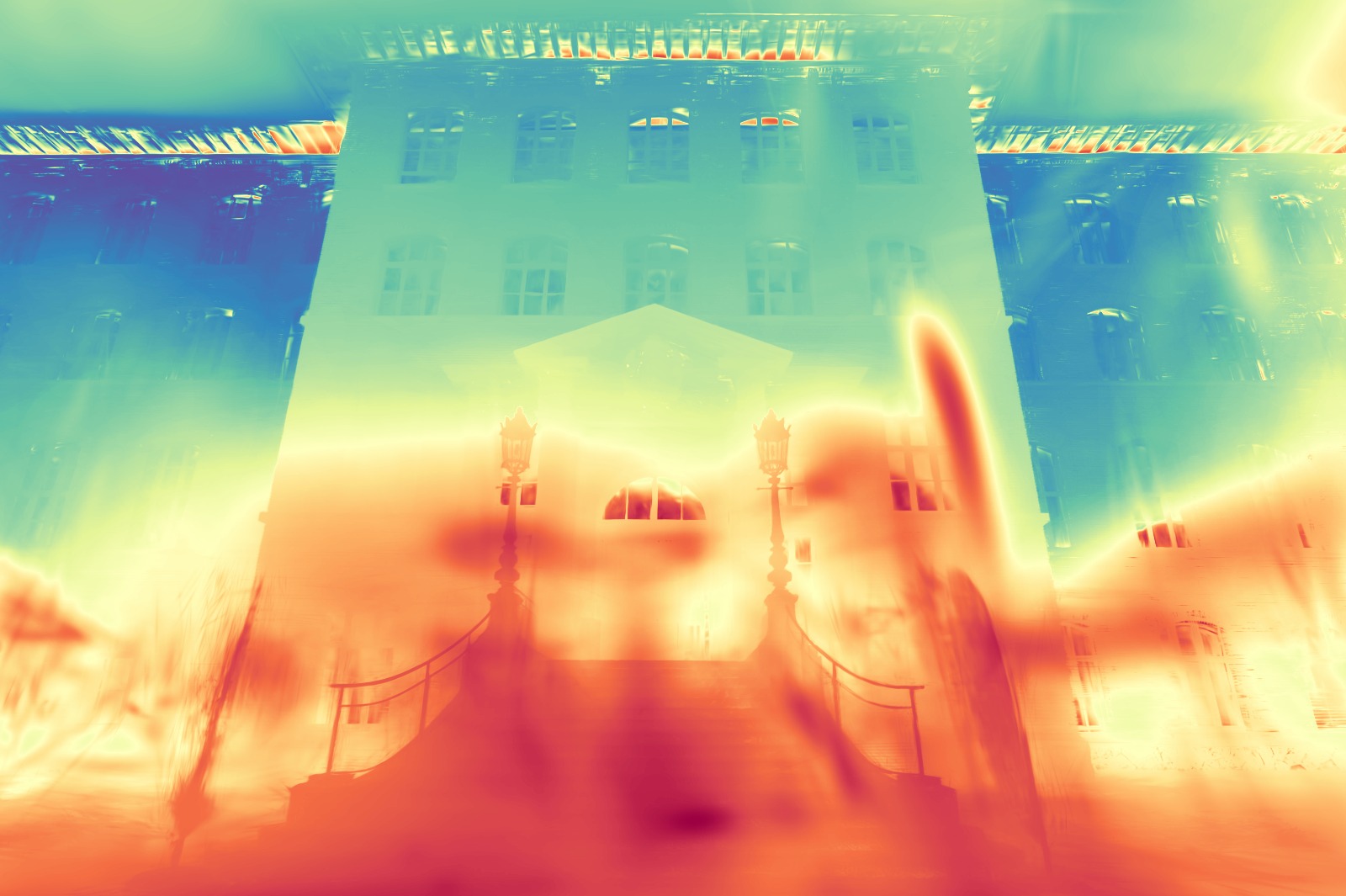} & \hspace{0.3cm} &
        \includegraphics[width=0.15\textwidth]{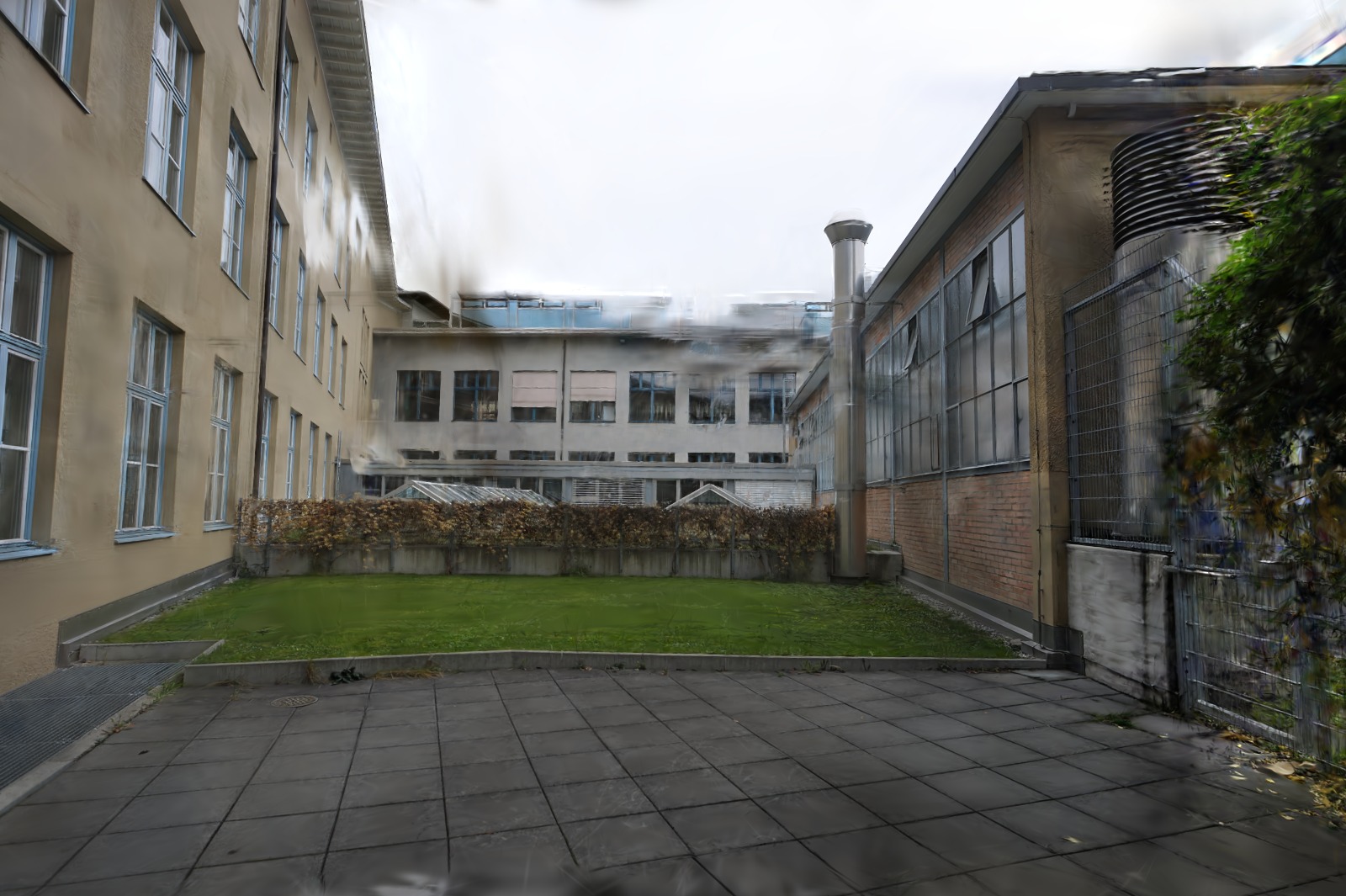} &
        \includegraphics[width=0.15\textwidth]{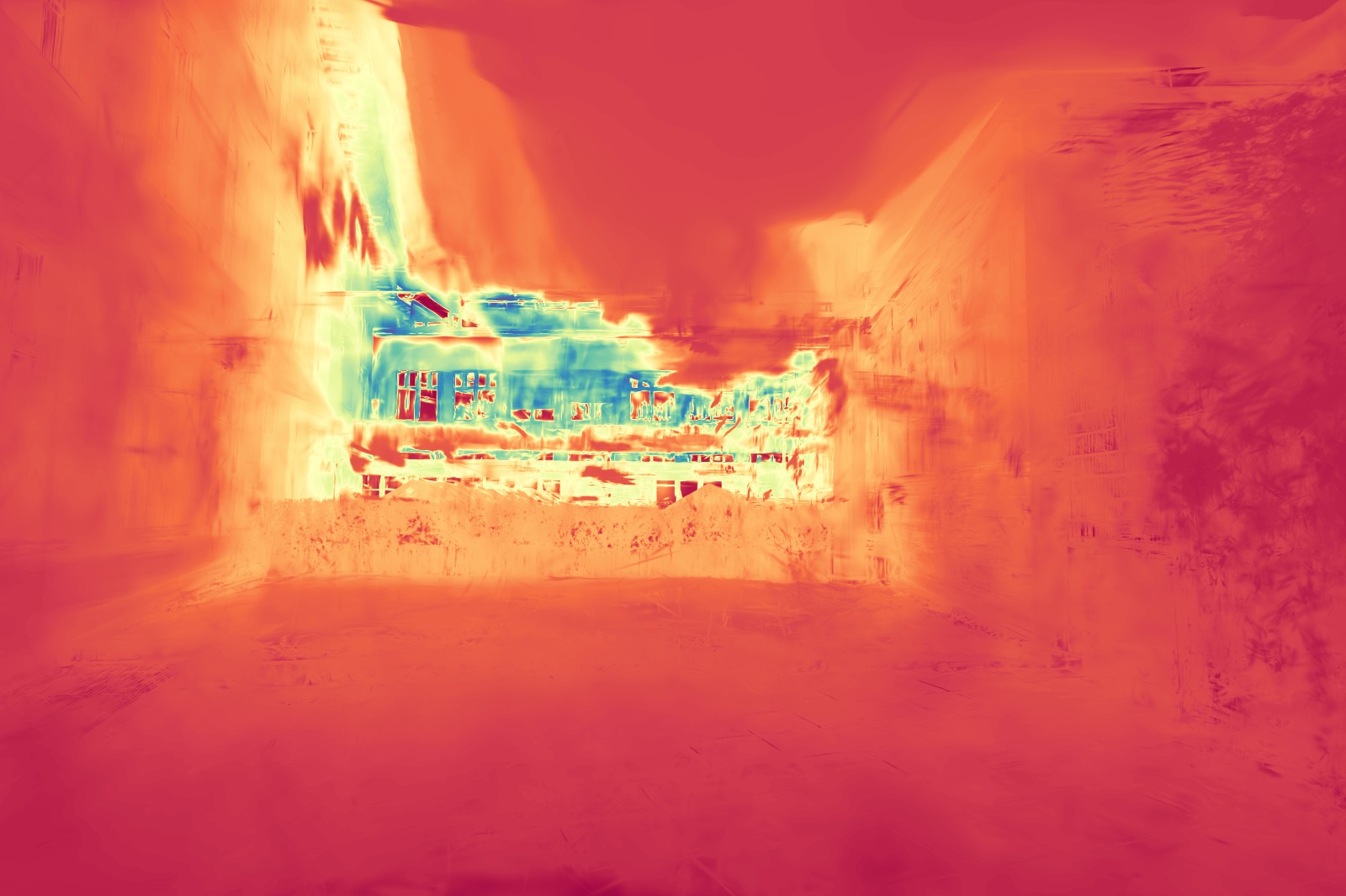} \\

    \rotatebox[origin=l]{90}{\scriptsize{\quad + SfM}} & 
        \includegraphics[width=0.15\textwidth]{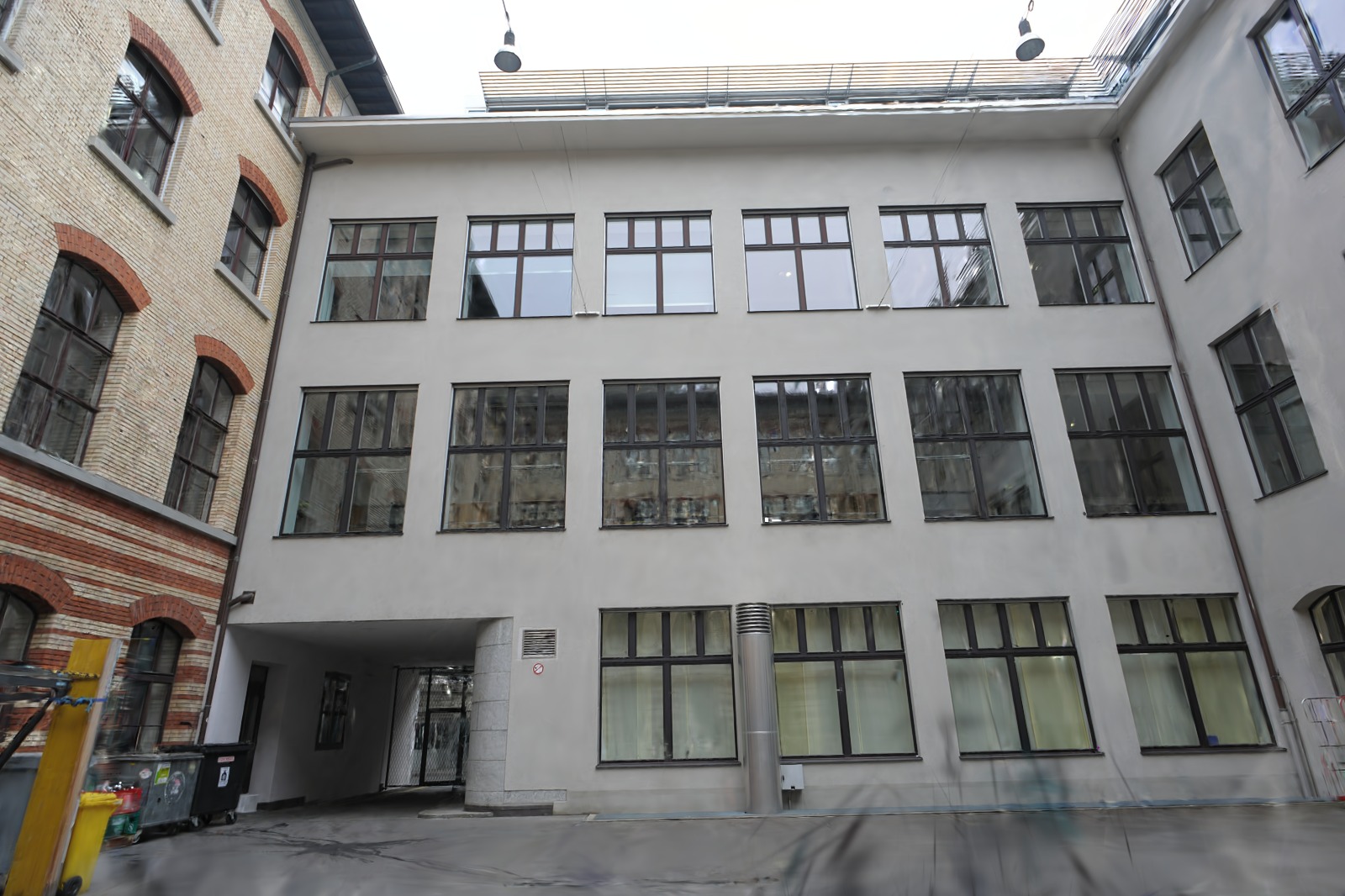} & 
        \includegraphics[width=0.15\textwidth]{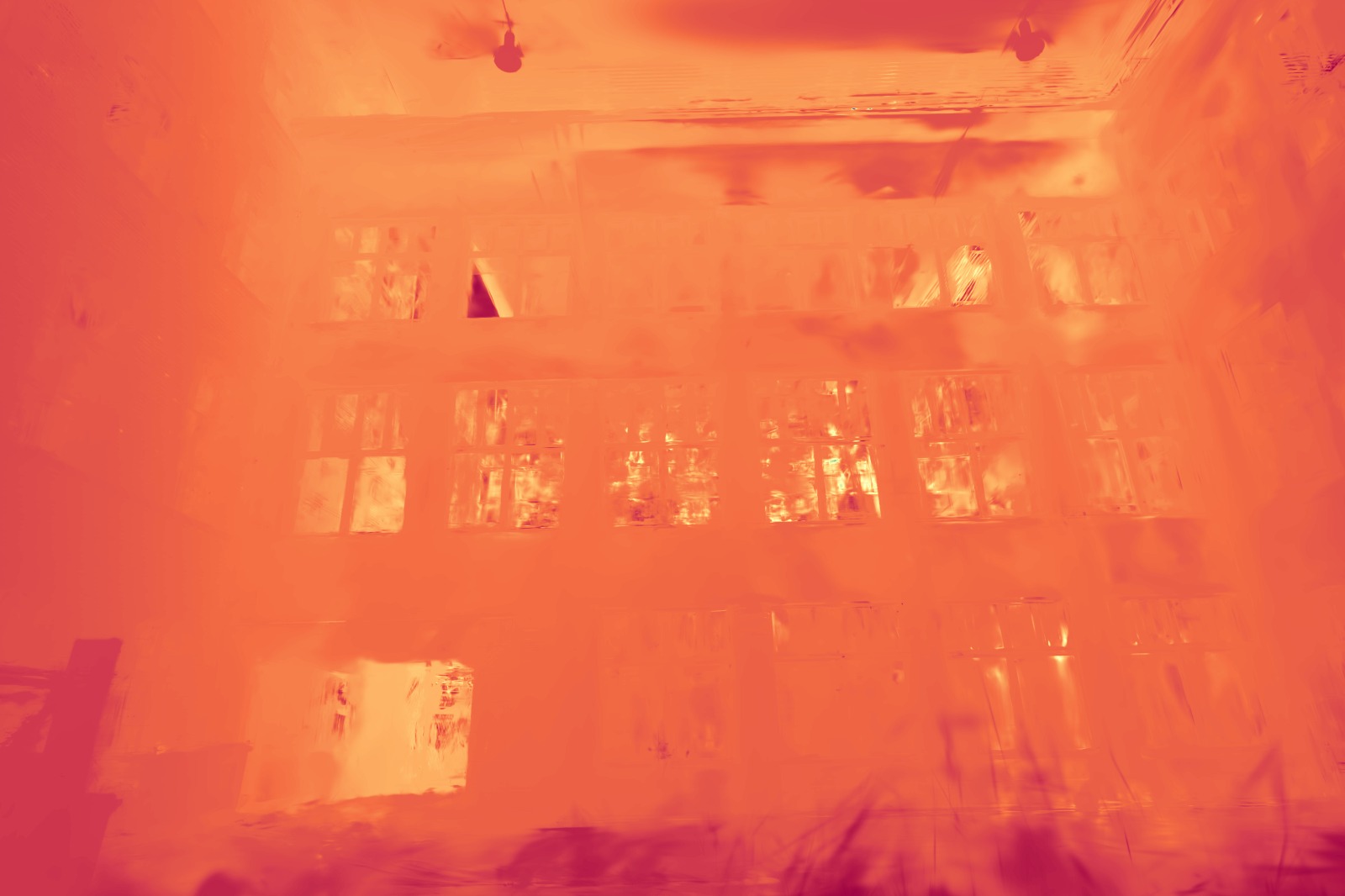} &\hspace{0.3cm} &
        \includegraphics[width=0.15\textwidth]{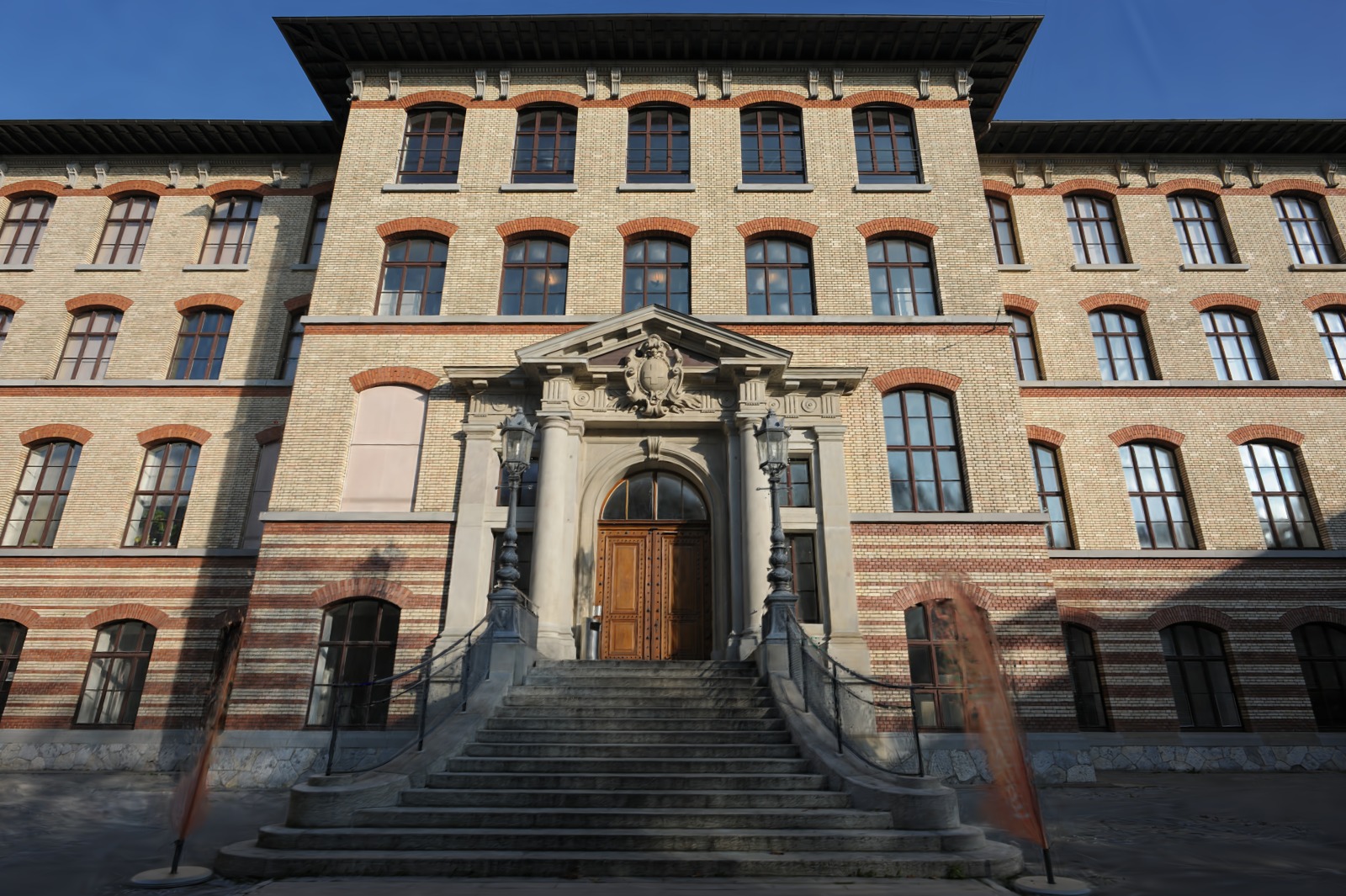} &
        \includegraphics[width=0.15\textwidth]{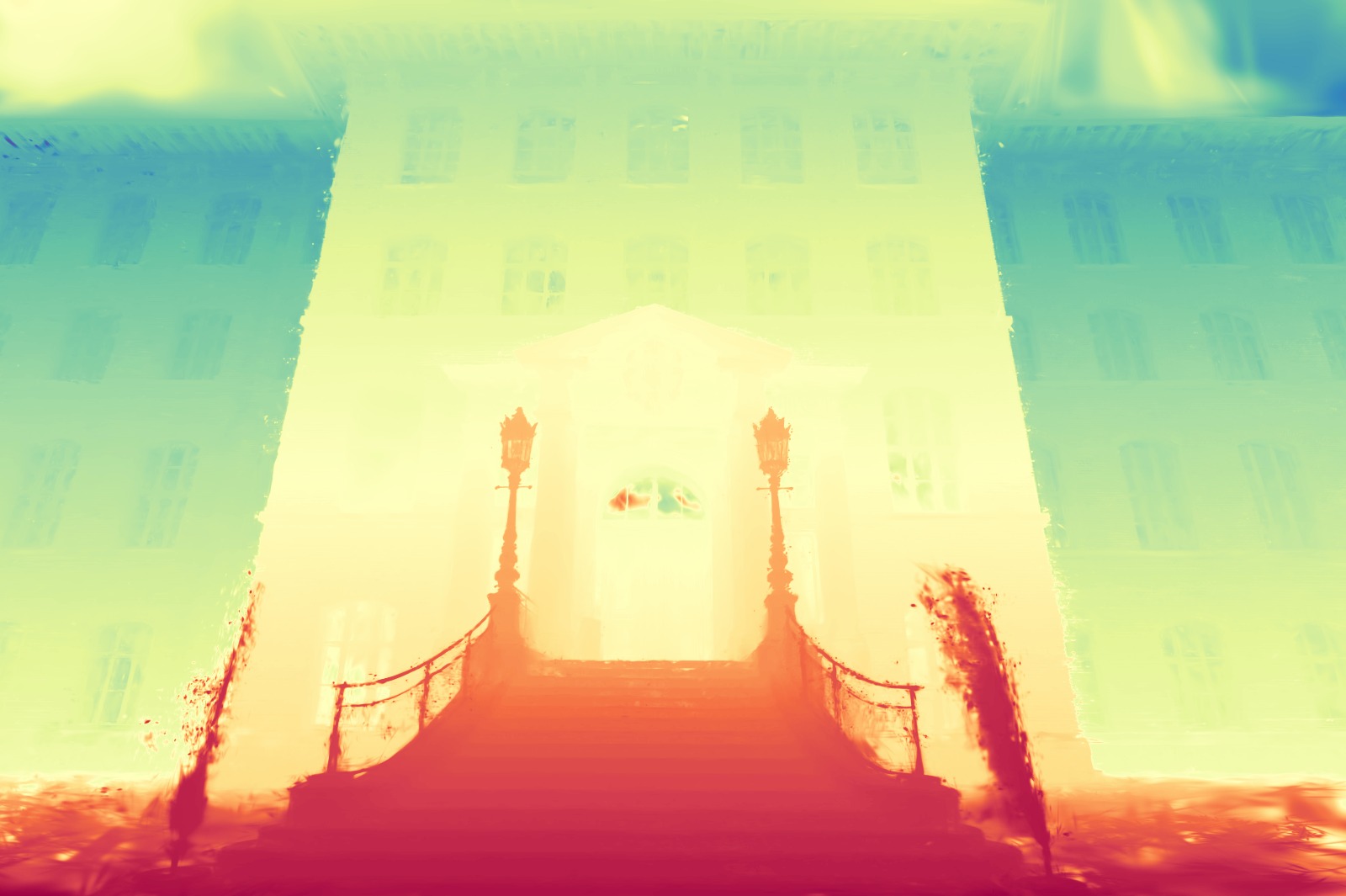} &\hspace{0.3cm} &
        \includegraphics[width=0.15\textwidth]{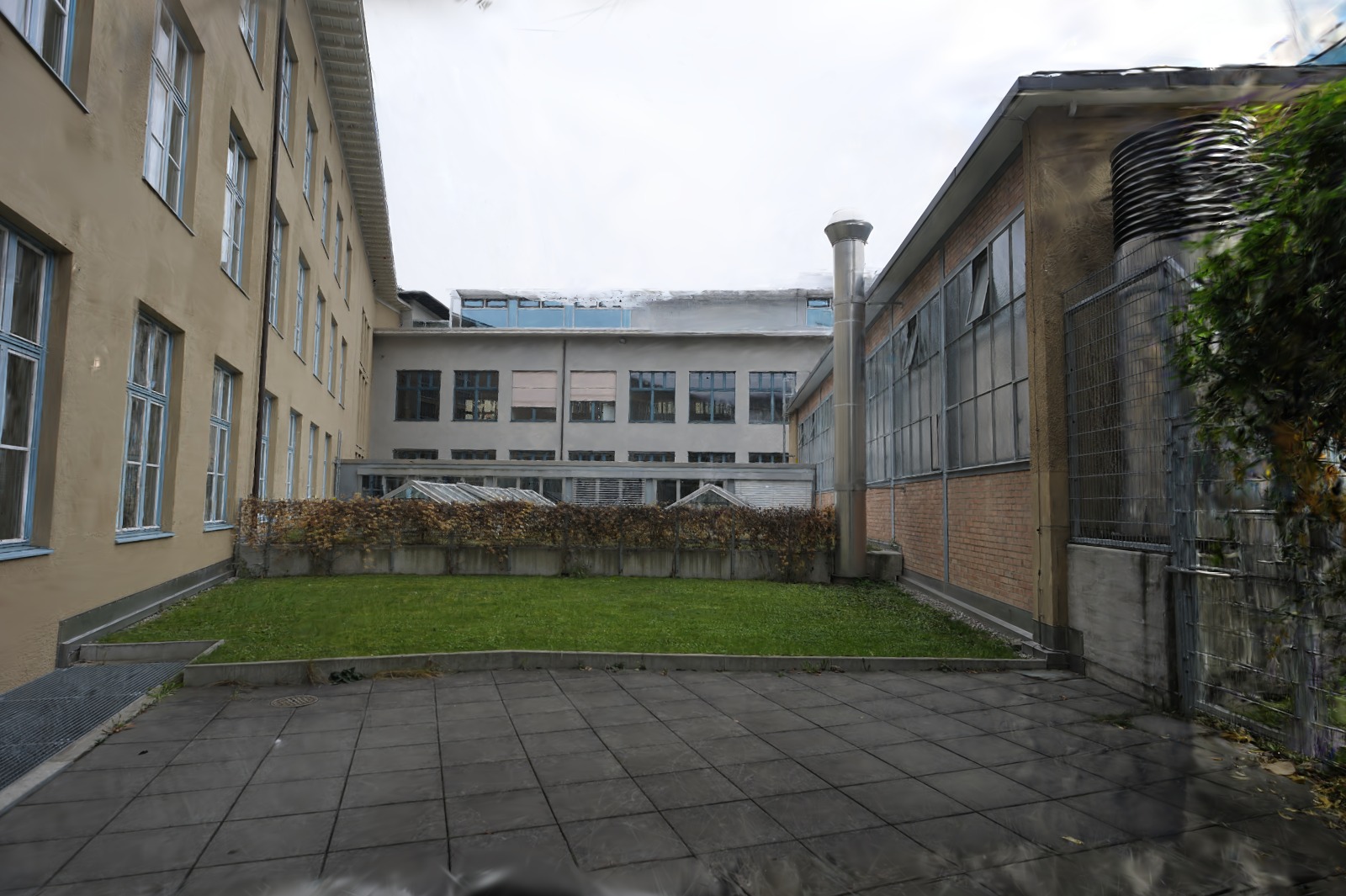} &
        \includegraphics[width=0.15\textwidth]{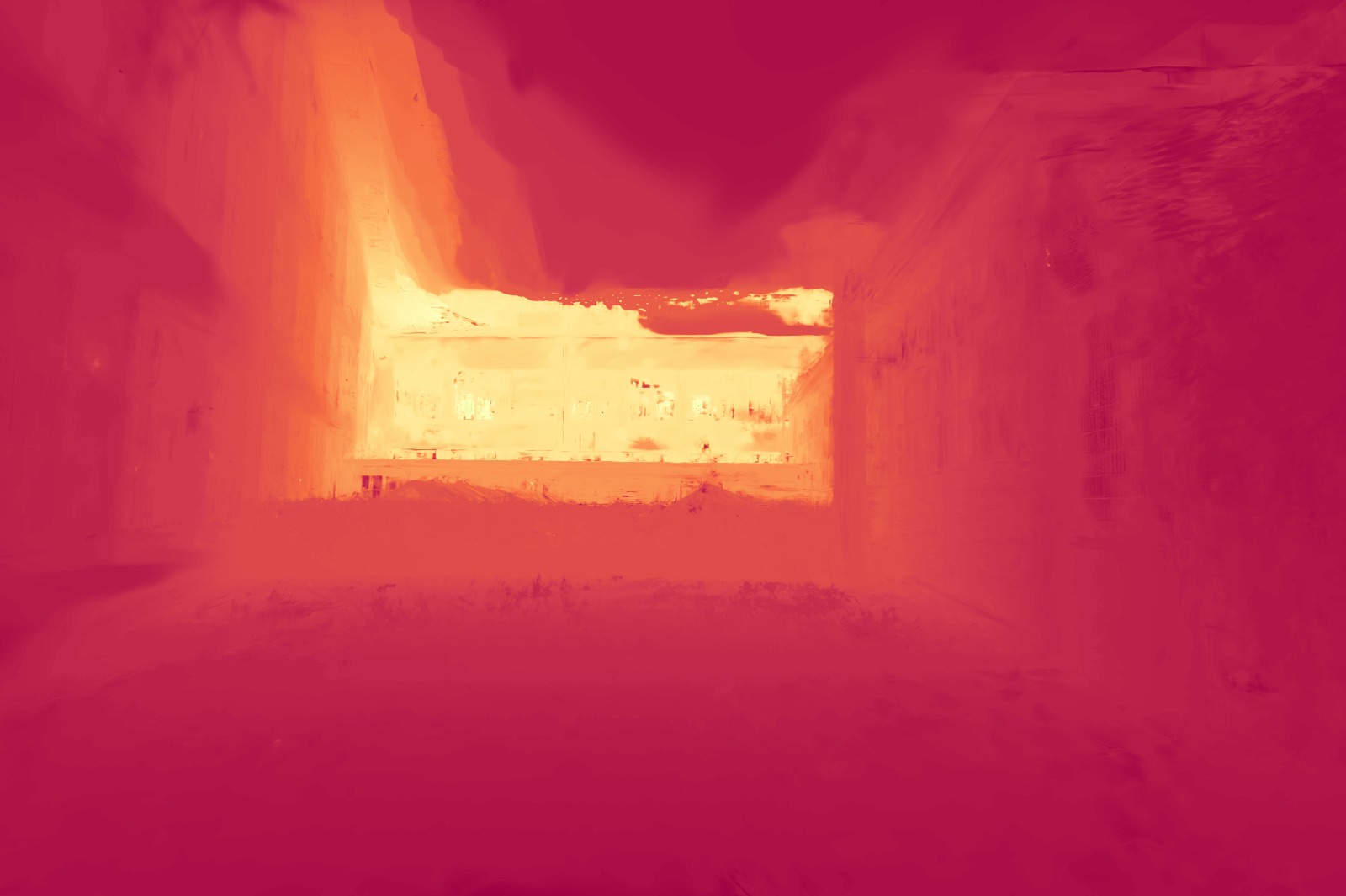} \\
        
        \rowcolor{First}
        \rotatebox[origin=l]{90}{\scriptsize{+ Self-Ev.}} & 
        \includegraphics[width=0.15\textwidth]{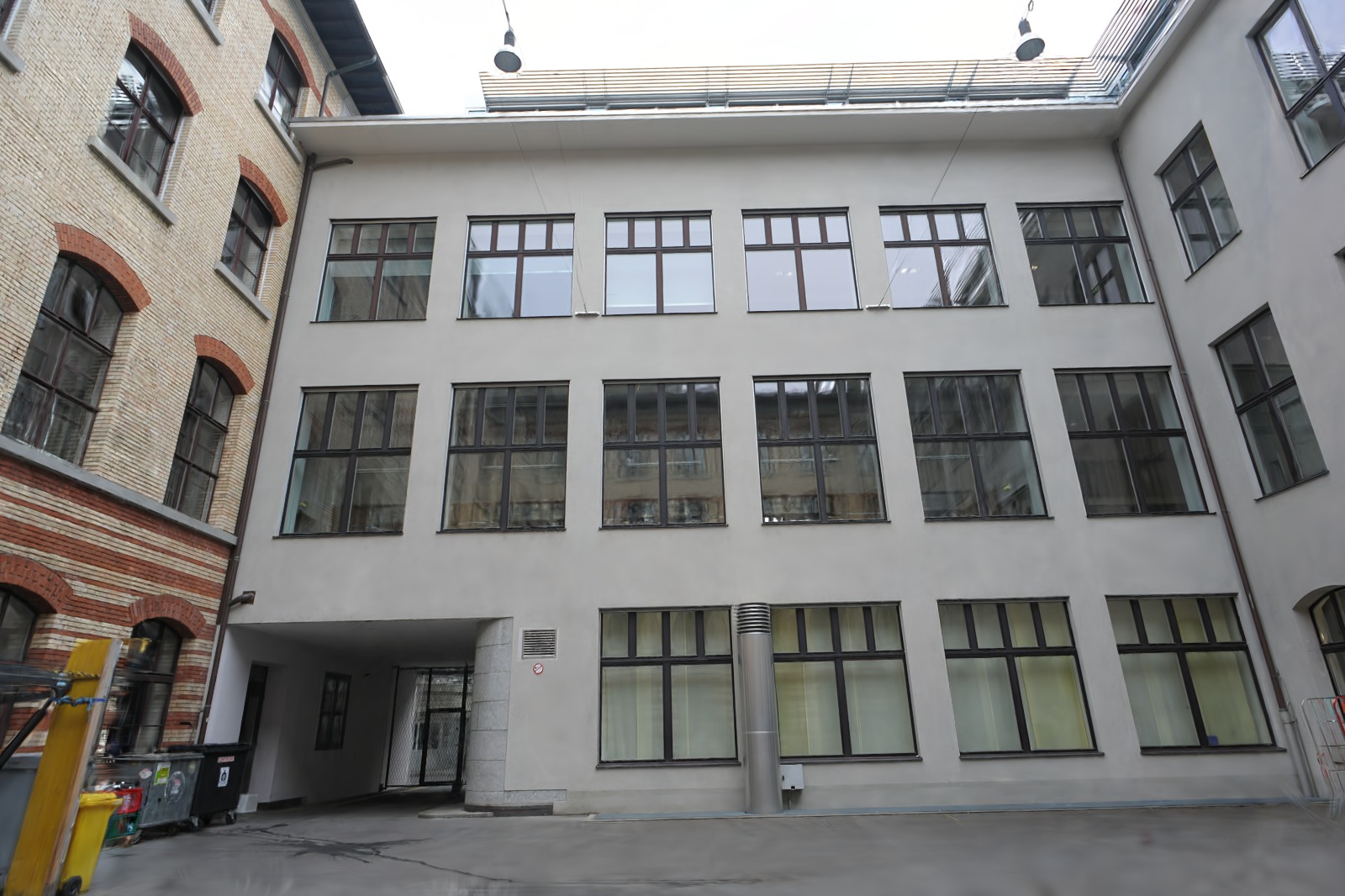} & 
        \includegraphics[width=0.15\textwidth]{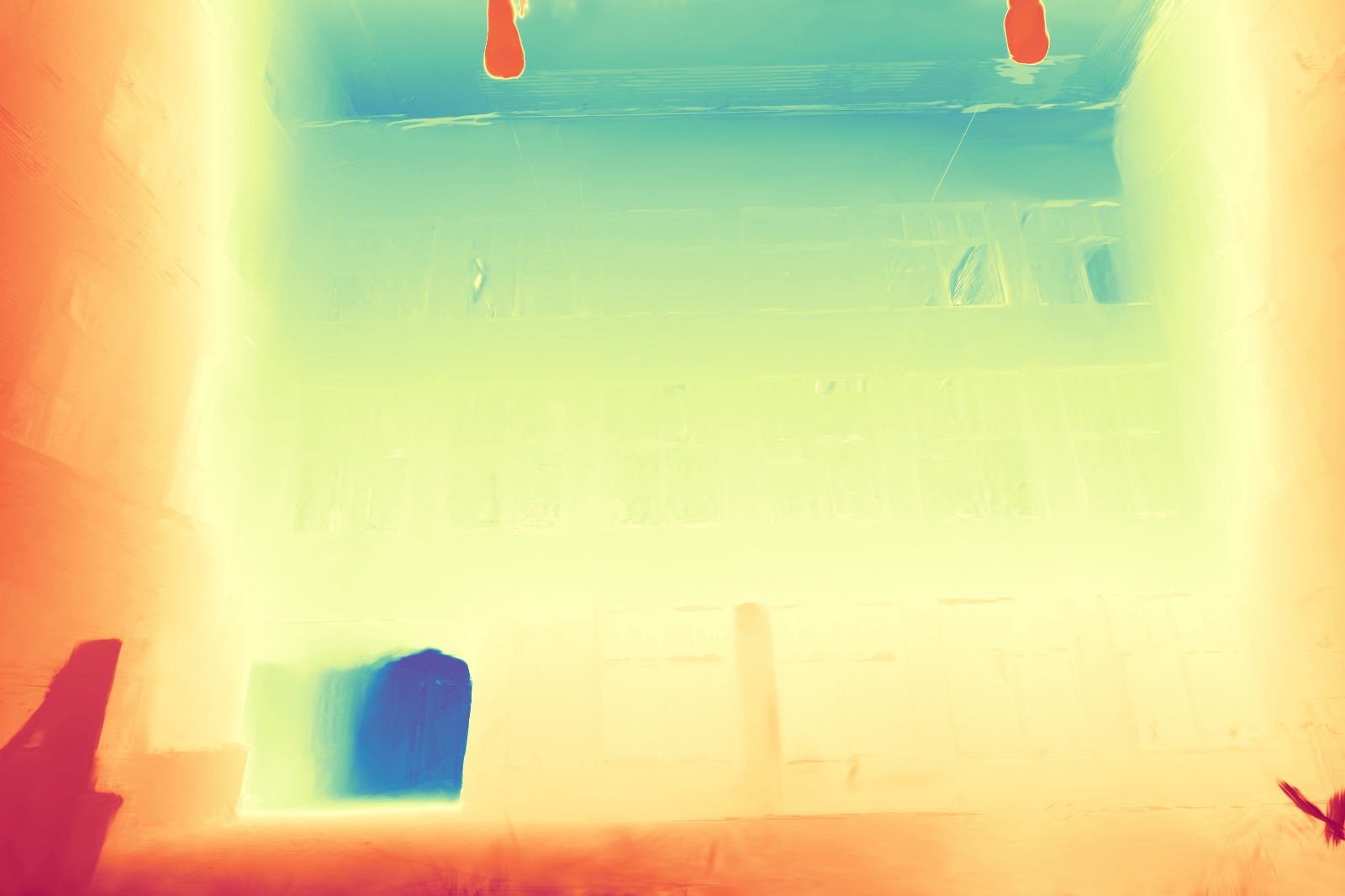} &\hspace{0.3cm} &
        \includegraphics[width=0.15\textwidth]{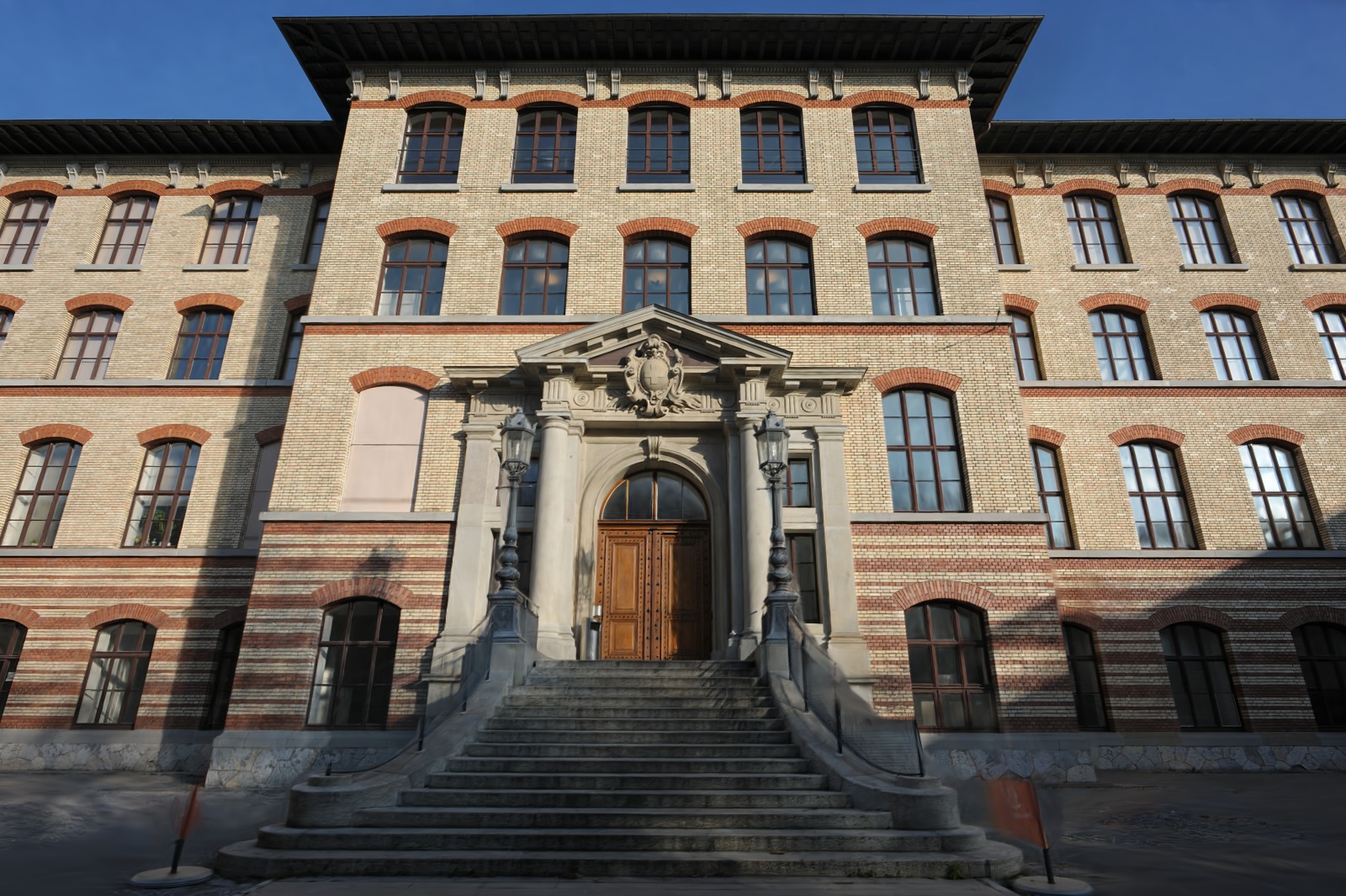} &
        \includegraphics[width=0.15\textwidth]{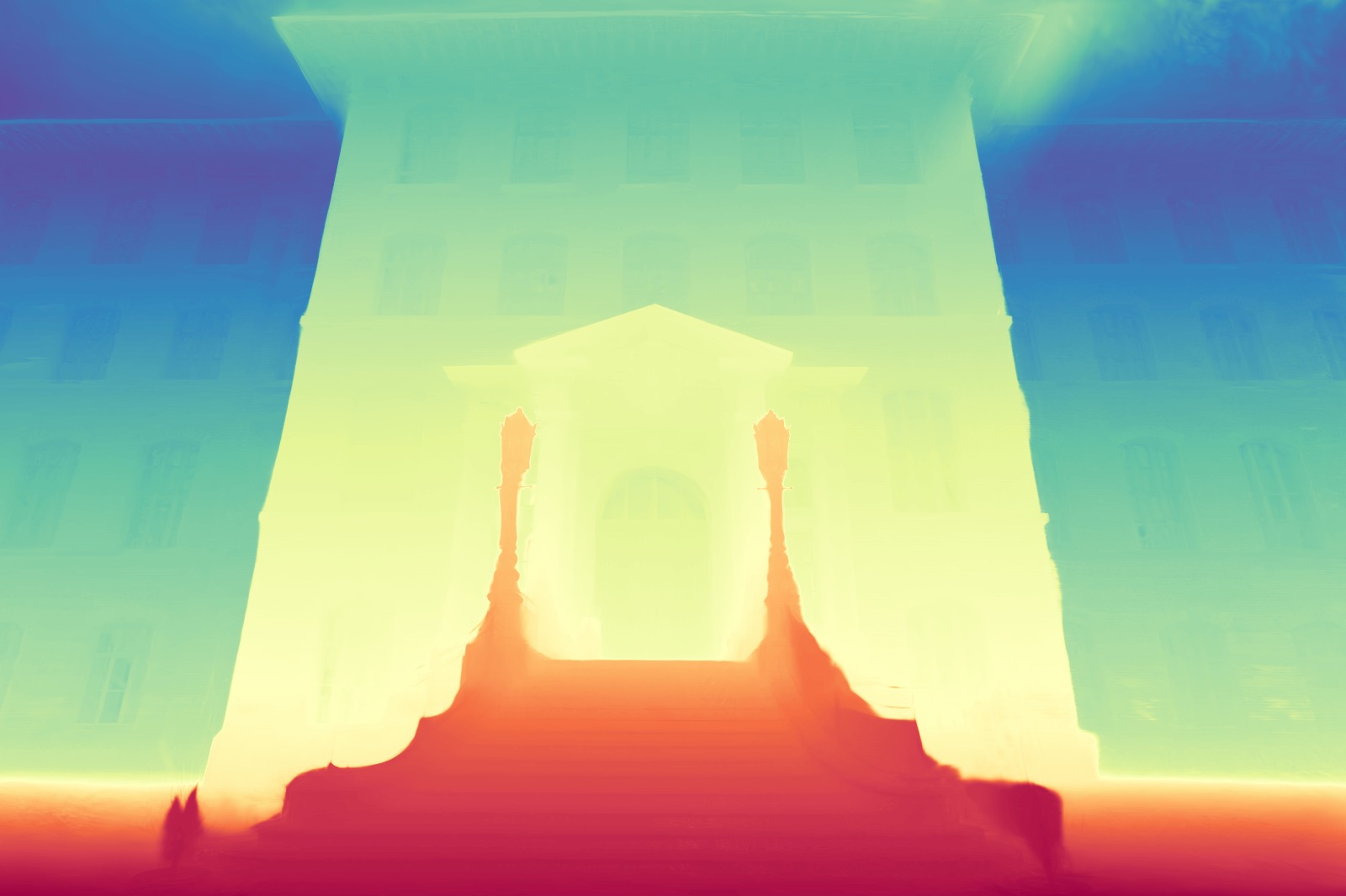} &\hspace{0.3cm} &
        \includegraphics[width=0.15\textwidth]{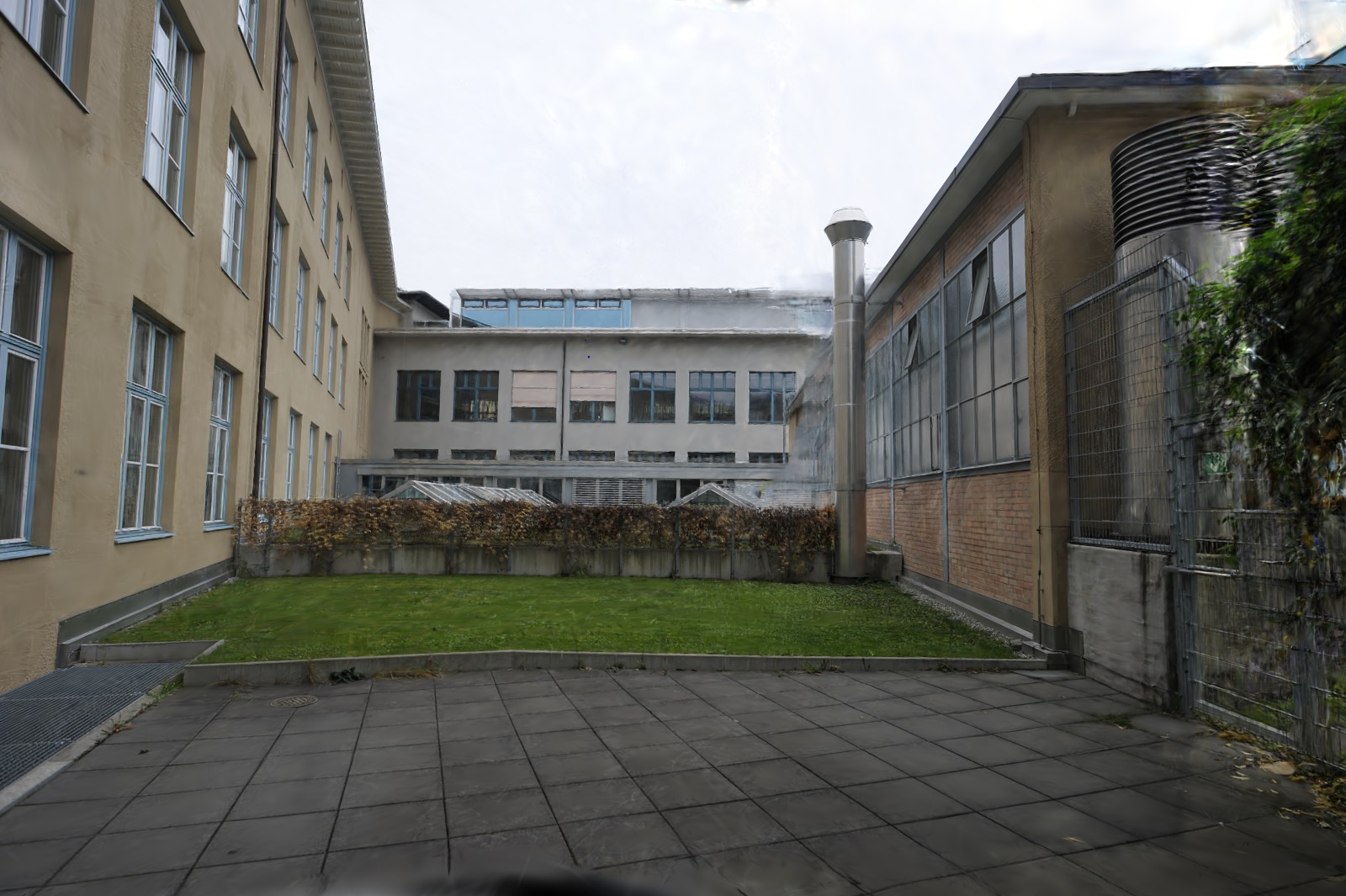} &
        \includegraphics[width=0.15\textwidth]{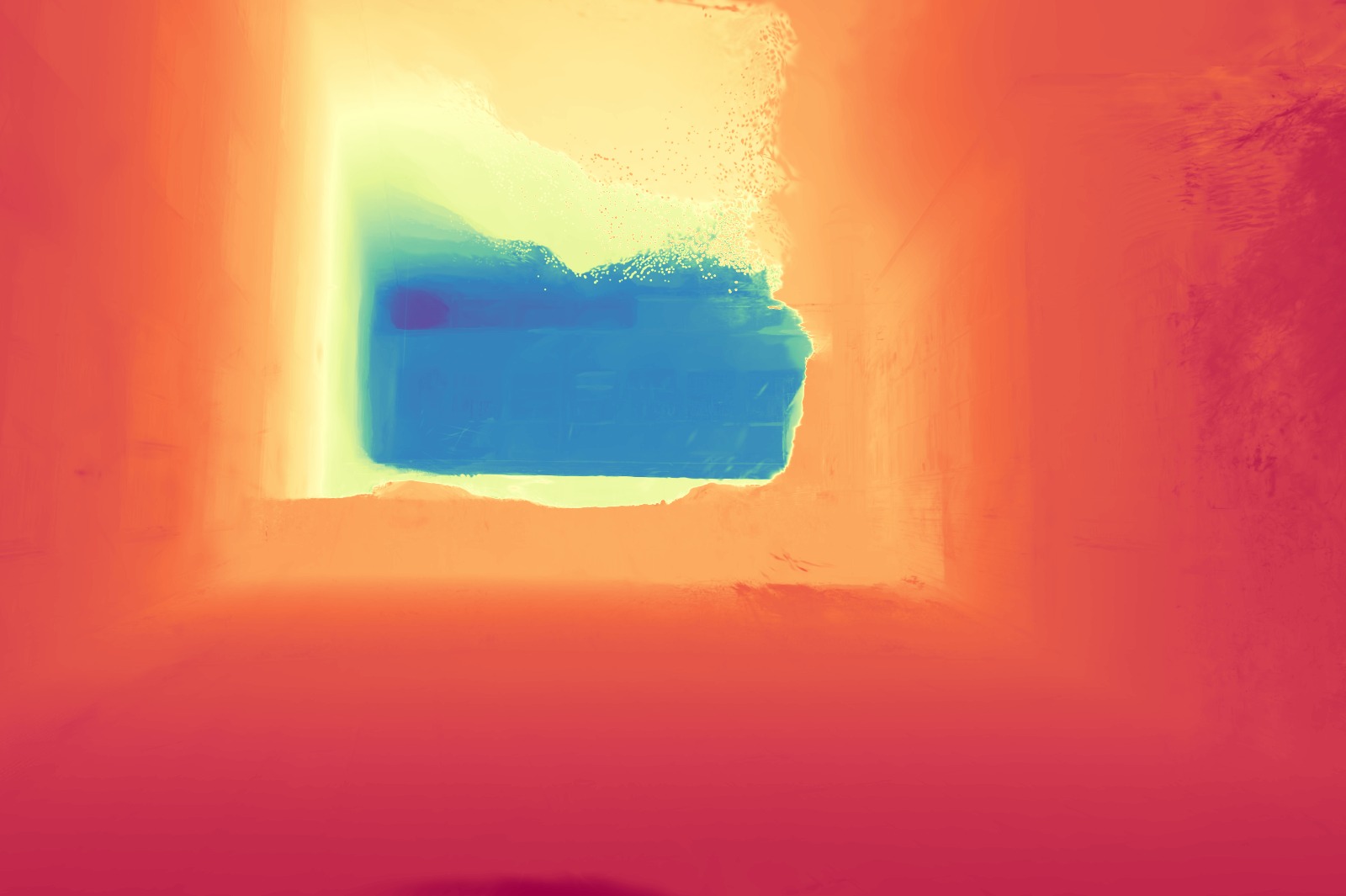}\vspace{0.1cm} \\

        \rotatebox[origin=l]{90}{\scriptsize{\quad GT}}
        &
        \includegraphics[width=0.15\textwidth]{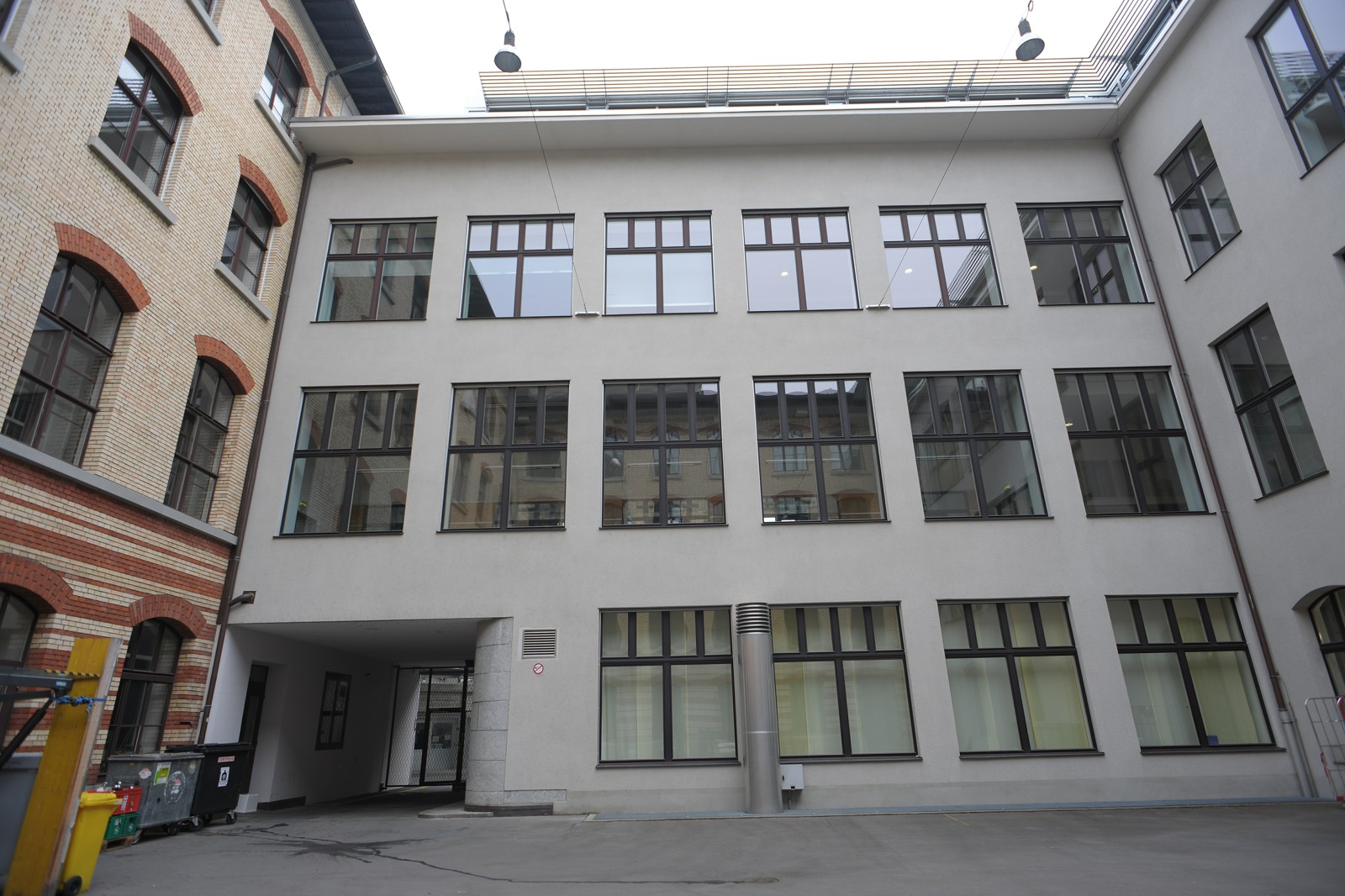} &
        \includegraphics[width=0.15\textwidth]{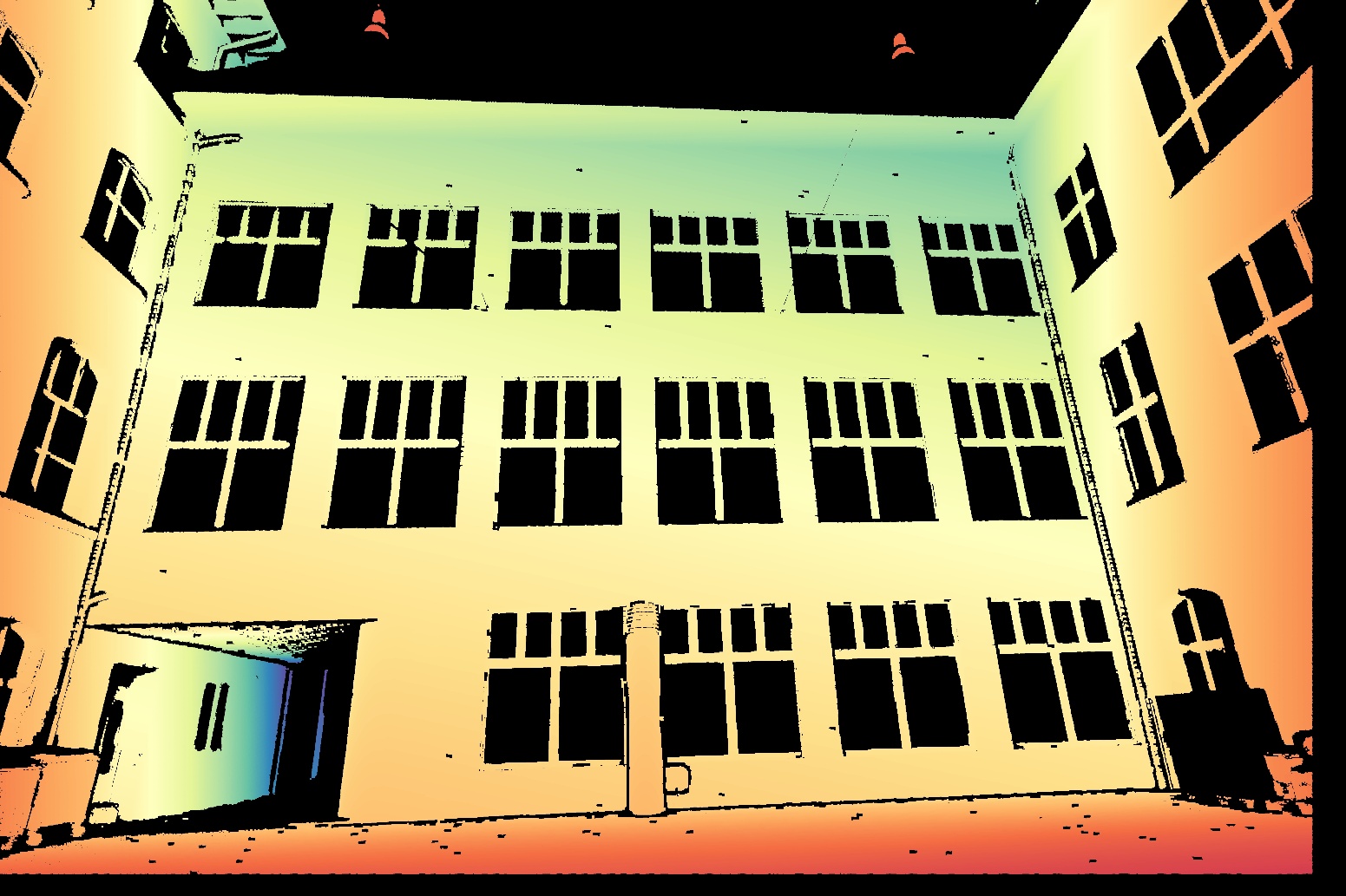} &\hspace{0.3cm} &
        \includegraphics[width=0.15\textwidth]{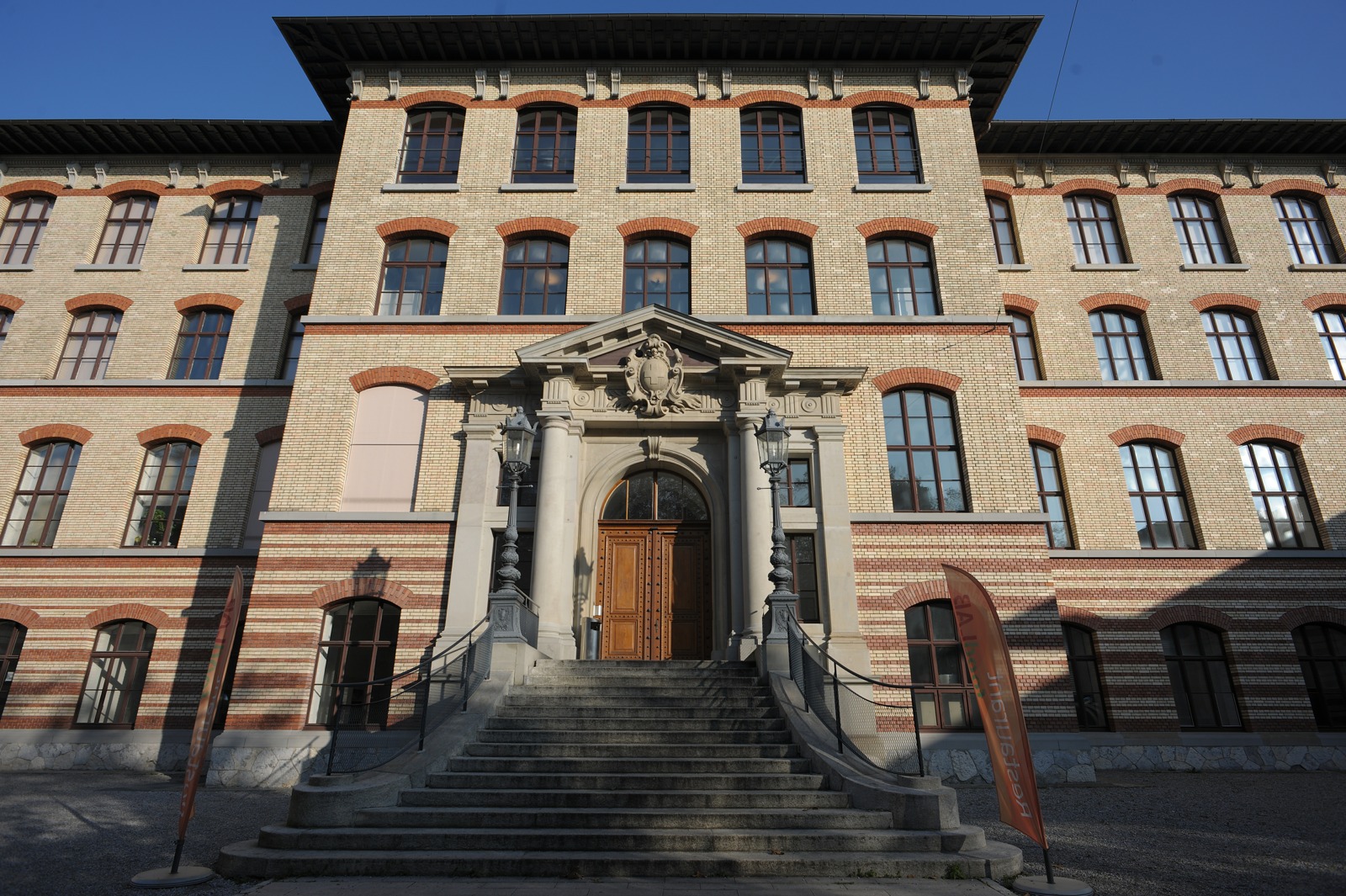} &
        \includegraphics[width=0.15\textwidth]{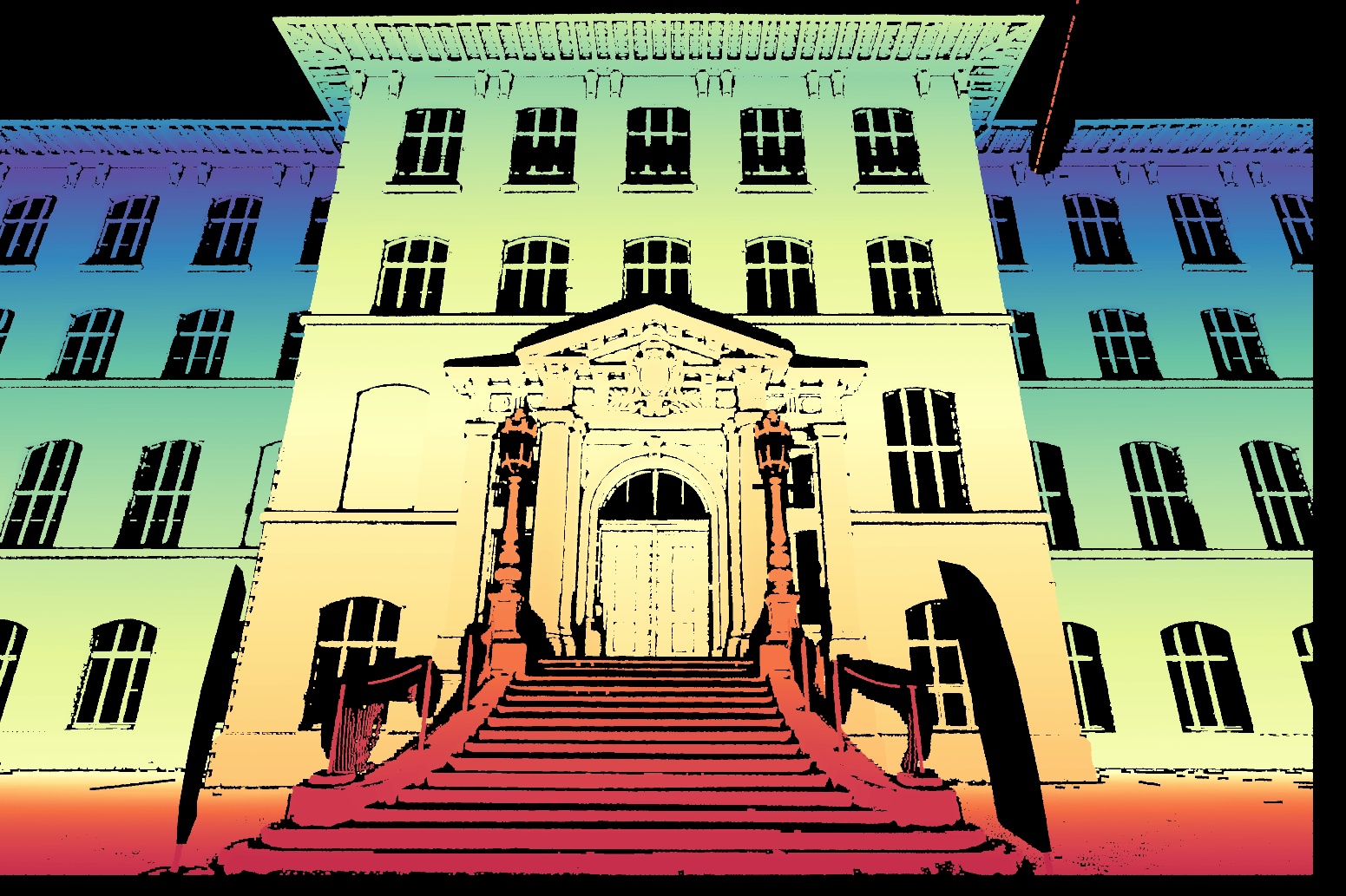} &\hspace{0.3cm} &
        \includegraphics[width=0.15\textwidth]{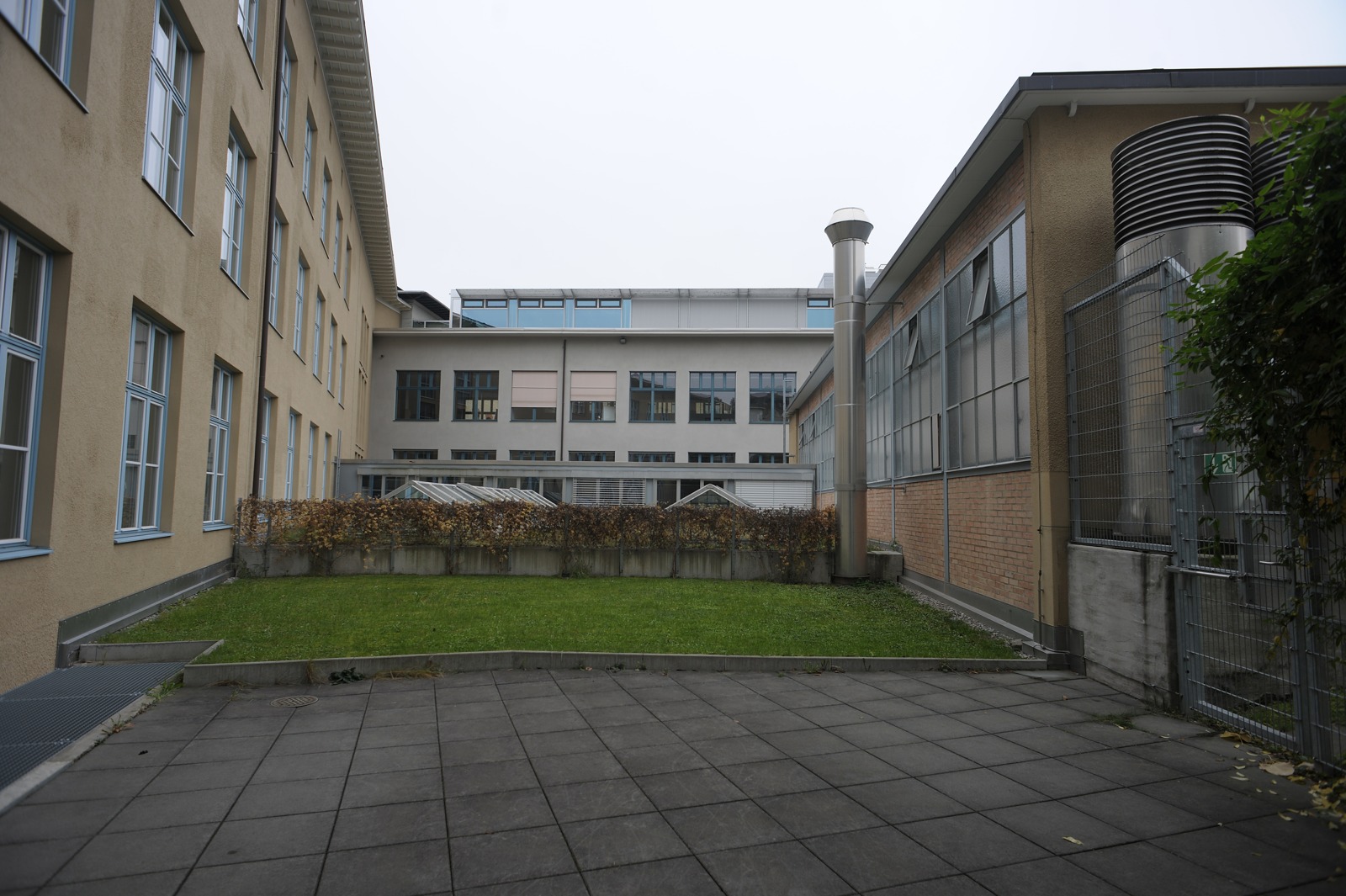} &
        \includegraphics[width=0.15\textwidth]{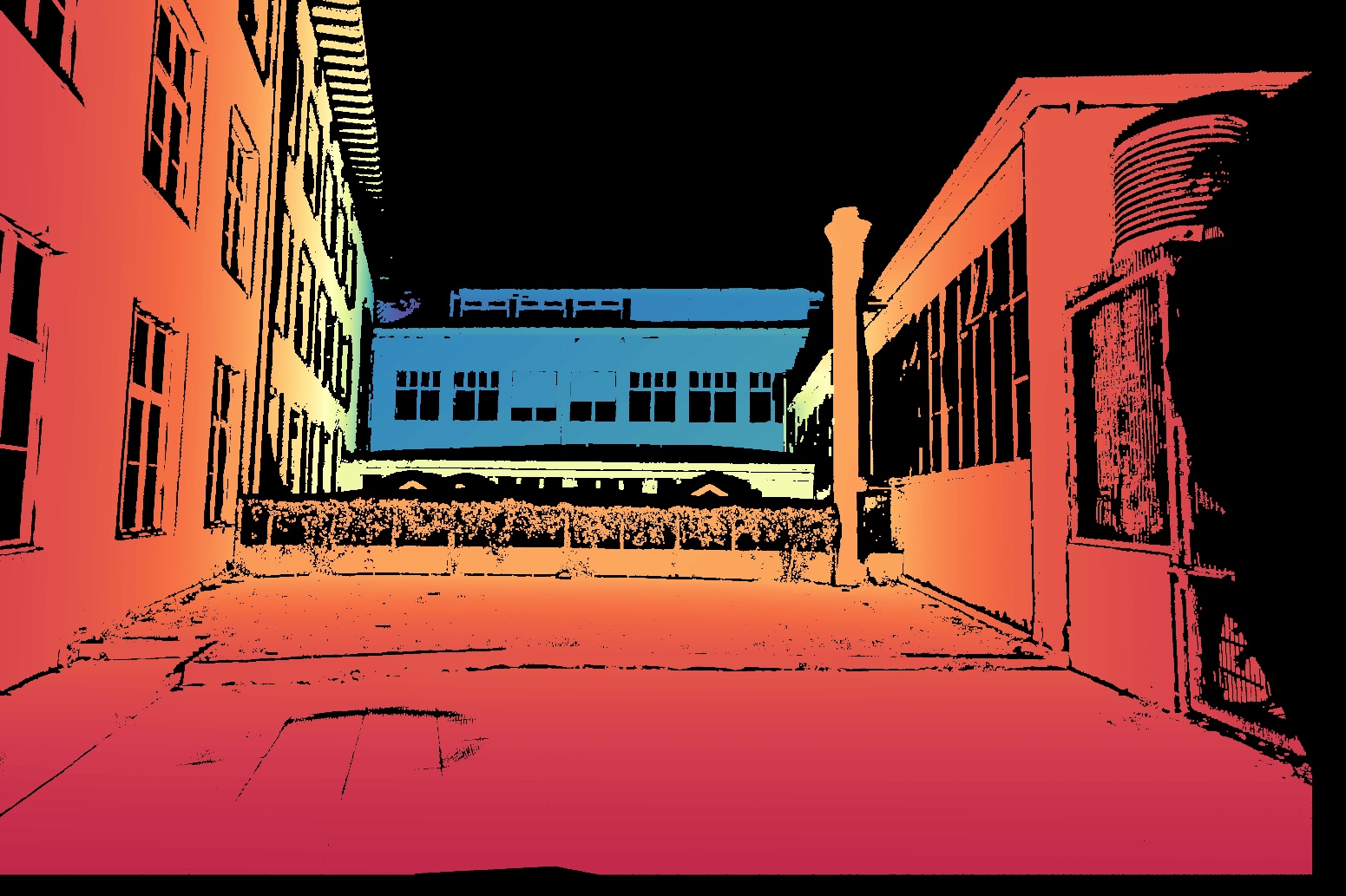} \\       
    \end{tabular}
    \caption{\textbf{Qualitative Results on ETH3D.}}
    \label{fig:qual_eth3d}
\end{figure*}

\begin{table}[t]
    \centering
\renewcommand{\tabcolsep}{10pt}
\begin{adjustbox}{max width=\textwidth}
    \begin{tabular}{cl ccc cccccc}
        & & \multicolumn{3}{c}{Depth} & \multicolumn{3}{c}{View Synthesis} \\
        \cmidrule(lr){1-2} \cmidrule(lr){3-5} \cmidrule(lr){6-8}
        & Method & Abs. Rel. $\downarrow$ & RMSE  $\downarrow$& $\delta < 1.25$  $\uparrow$ & SSIM  $\uparrow$& PSNR  $\uparrow$& LPIPS  $\downarrow$  \\
        \cmidrule(lr){1-2} \cmidrule(lr){3-5} \cmidrule(lr){6-8}
        & GS \cite{kerbl20233d} & 0.058 & 7.041 & 0.933 & 0.6301 & 21.6160 & 0.2729   \\

        \addlinespace
        \hdashline
        \addlinespace
        \multirow{4}{*}{\rotatebox[origin=l]{90}{{+ Self-Ev.}}} & IGEV-Stereo \cite{hamzah2016literature} (Middlebury \cite{scharstein2014high}) & 0.025 & 4.331 & 0.984 & 0.6340 & 21.8255 & 0.2727   \\
        & PCVNet \cite{zeng2023parameterized} (SceneFlow \cite{mayer2016large}) & \trd 0.023 & \trd 4.082 & \snd 0.988 &  \trd 0.6351 & \trd 21.8605 & \trd 0.2717  \\
        & CREStereo \cite{li2022practical} (Mixed) & \snd 0.023 & \snd 3.896 & \trd 0.985 & \snd 0.6361 & \snd 21.9168 & \snd 0.2714  \\
        & iRaftStereo\_RVC \cite{jiang2022iraft} (Mixed) & \fst 0.020 & \fst 3.714 & \fst 0.992 & \fst 0.6377 & \fst 21.9734 & \fst 0.2696  \\
        \hline
    \end{tabular}
    \end{adjustbox}
    \caption{\textbf{Ablation Studies on BlendedMVS -- stereo models.}}
    \label{tab:ablation_models}
\end{table}

\begin{table}[t]
    \centering
\renewcommand{\tabcolsep}{10pt}
\begin{adjustbox}{max width=\textwidth}
    \begin{tabular}{cl ccc cccccc}
        & & \multicolumn{3}{c}{Depth} & \multicolumn{3}{c}{View Synthesis} \\
        \cmidrule(lr){1-2} \cmidrule(lr){3-5} \cmidrule(lr){6-8}
        & Method & Abs. Rel. $\downarrow$ & RMSE  $\downarrow$& $\delta < 1.25$  $\uparrow$ & SSIM  $\uparrow$& PSNR  $\uparrow$& LPIPS  $\downarrow$ \\
        \cmidrule(lr){1-2} \cmidrule(lr){3-5} \cmidrule(lr){6-8}
        & GS \cite{kerbl20233d} & 0.058 & 7.041 & 0.933 & 0.6301 & 21.6160 & 0.2729   \\

        \addlinespace
        \hdashline
        \addlinespace
        \multirow{4}{*}{\rotatebox[origin=l]{90}{+ Self-Ev.}} & RAFT-Stereo \cite{lipson2021raft} (NerfStereo \cite{Tosi_2023_CVPR}) & 0.023 & 4.046 & 0.987 & 0.6360 & 21.8964 & 0.2715   \\
        & RAFT-Stereo \cite{lipson2021raft} (Sceneflow \cite{mayer2016large}) & \fst 0.020 & \fst 3.679 & \snd 0.991 & \snd 0.6375 & \snd 21.9517 & \fst 0.2696  \\
        & RAFT-Stereo \cite{lipson2021raft} (Middlebury \cite{scharstein2014high}) & \trd 0.021 & \trd 3.929 & \trd 0.989 & \trd 0.6362 & \trd 21.9027 & \trd 0.2708 \\
        & iRaftStereo\_RVC \cite{jiang2022iraft} (Mixed) & \fst 0.020 & \snd 3.714 & \fst 0.992 & \fst 0.6377 & \fst 21.9734 & \fst 0.2696  \\

        \hline

    \end{tabular}
    \end{adjustbox}
    \caption{\textbf{Ablation Studies on BlendedMVS -- RAFT-Stereo variants.}}\label{tab:ablation_raft}
\end{table}

\subsection{Ablation Studies}

Finally, we conduct some ablation studies focused on our self-evolving framework. These are carried out on the BlendedMVS dataset and averaged over 5 runs. 

\textbf{Stereo Models.} \cref{tab:ablation_models} collects the results achieved by using different stereo networks to obtain depth priors. For each model, we report the specific weights we used among those available in brackets. We can appreciate how using any of the state-of-the-art stereo backbones allows for largely improving the results over vanilla GS. However, iRAFTStereo\_RCV and CREStereo show higher improvements against PCVNet and IGEV-Stereo, both in terms of color and depth rendering. We ascribe this both to their specific architecture, as well as to the mix of several datasets used to train them.

\textbf{RAFT-Stereo -- Training Datasets.} To figure out the real impact of both the training data and the stereo backbone, in \cref{tab:ablation_raft} we compare the results obtained with different RAFT-Stereo weights. The gap between the several instances is very low, with iRAFT-Stereo and the original RAFT-Stereo trained on SceneFlow being on par on three out of six metrics.

\section{Conclusion} 
\label{sec:conclusion}

In summary, this work seeks to address a critical limitation in 3D Gaussian Splatting by focusing on improving its underlying scene geometry. Through a comprehensive analysis, we study an optimization approach that integrates external depth priors, simultaneously improving the inferred 3D structure and the quality of novel view synthesis. A key contribution is our novel strategy leveraging depth priors computed from readily available deep stereo networks on virtual stereo pairs rendered during training by GS itself, demonstrating superior performance compared to alternative depth-from-image solutions. Experimental results on ETH3D, ScanNet++, and BlendedMVS datasets support the importance of our findings and provide evidence for the effectiveness of our proposal.

\textbf{Acknowledgements.} We acknowledge the CINECA award under the ISCRA initiative, for the availability of high-performance computing resources and support. Sadra Safadoust was supported by
KUIS AI Fellowship and UNVEST R\&D Center. This project is co-funded by the European Union (ERC, ENSURE, 101116486). Views and opinions expressed are however those of the author(s) only and do not necessarily reflect those of the European Union or the European Research Council. Neither the European Union nor the granting authority can be held responsible for them.

\bibliography{main}
\end{document}